\documentclass[sigconf, nonacm]{acmart}

\usepackage{pvldb}
\usepackage{float}
\usepackage{booktabs}
\usepackage{multirow}
\usepackage{graphicx}
\usepackage{array,colortbl,tabularx}
\definecolor{cmpOursBg}{RGB}{245,245,220}
\definecolor{cmpGainBg}{RGB}{255,221,221}
\usepackage{xcolor}

\usepackage{bigstrut}

        \usepackage{tikz}

\usepackage{xcolor}
\usepackage{tikz}

\newcommand{\italiccircled}[2][red]{%
    \tikz[baseline=(char.base)]
    \node[
        draw=#1,
        fill=#1!12,
        text=#1,
        circle,
        inner sep=0.8pt,
        line width=0.5pt,
        font=\bfseries\itshape
    ] (char) {#2};%
}
\microtypesetup{protrusion=false}
\renewcommand\vldbavailabilityurl{https://github.com/JunxinLu001/ReFi-TS}

\begin{document}
\title{Temporal Relational Fingerprint-Guided Normality Learning for Time-Series Anomaly Detection}

\author{Junxin Lu}
\affiliation{
  \institution{School of Communication and Information Engineering,
               Shanghai University}
  \city{Shanghai}
  \country{China}
}
\email{junxinlu@shu.edu.cn}

\author{Siyan Zhu}
\affiliation{
  \institution{School of Communication and Information Engineering,
               Shanghai University}
  \city{Shanghai}
  \country{China}
}
\email{zsy18868751048@gmail.com}

\author{Xingjian Tian}
\affiliation{
  \institution{School of Communication and Information Engineering,
               Shanghai University}
  \city{Shanghai}
  \country{China}
}
\email{tianxingjian@shu.edu.cn}

\author{Zhijun Fang}
\affiliation{
  \institution{Institute of Trustworthy Embodied AI, Fudan University}
  \city{Shanghai}
  \country{China}
}
\email{zjfang@fudan.edu.cn}

\author{Shiliang Sun}
\authornote{Corresponding authors.}
\affiliation{
  \institution{School of Automation and Intelligent Sensing,
               Shanghai Jiao Tong University}
  \city{Shanghai}
  \country{China}
}
\email{shiliangsun@gmail.com}
 
\author{Dan Zeng}
\authornotemark[1]
\affiliation{
  \institution{School of Communication and Information Engineering,
               Shanghai University}
  \city{Shanghai}
  \country{China}
}
\email{dzeng@shu.edu.cn}

\begin{abstract}
Effective time series anomaly detection (TSAD) relies on learning 
discriminative temporal representations that capture the 
underlying regularities of normal behaviors.
However, existing representation-based 
methods predominantly rely on implicit feature similarities,  which may fail to preserve 
contextual statistical relationships and intrinsic temporal structures.
This obscures subtle anomaly patterns and hinders the
characterization of heterogeneous normal behaviors.
To address these limitations, we propose ReFi-TS, a novel method 
that integrates explicit temporal relational statistical semantics into deep 
representation learning for TSAD.
Specifically,  ReFi-TS constructs a five-dimensional Temporal Relational Fingerprint (ReFi)
to characterize contextual
statistical variations and intrinsic temporal structures for each patch.
A bidirectional cross-view alignment mechanism is introduced to 
preserve these statistical semantics in deep patch representations.
Furthermore,  a ReFi-guided Comparator
establishes correspondence between deep patch representations
and statistical normal modes, while soft mode-conditional hypersphere learning constrains mode-specific representation
distributions. 
During inference,
ReFi-TS derives complementary
anomaly evidence from deep representation discrepancy, global ReFi
rarity, and local ReFi disaffinity, which is integrated through a
two-stage fusion strategy. 
Extensive experiments on TSB-AD-U and TSB-AD-M demonstrate that 
ReFi-TS achieves the best performance across all evaluation metrics, improving VUS-PR over
the strongest baselines by \textbf{7$\%$} and \textbf{3$\%$}, respectively,
with only approximately \textbf{0.4M}  trainable parameters. 
Code is available at \url{https://github.com/JunxinLu001/ReFi-TS}.
\end{abstract}

\maketitle

\vldbtopmatter
\begin{figure}[]
\centering
\includegraphics[width=0.95\columnwidth]{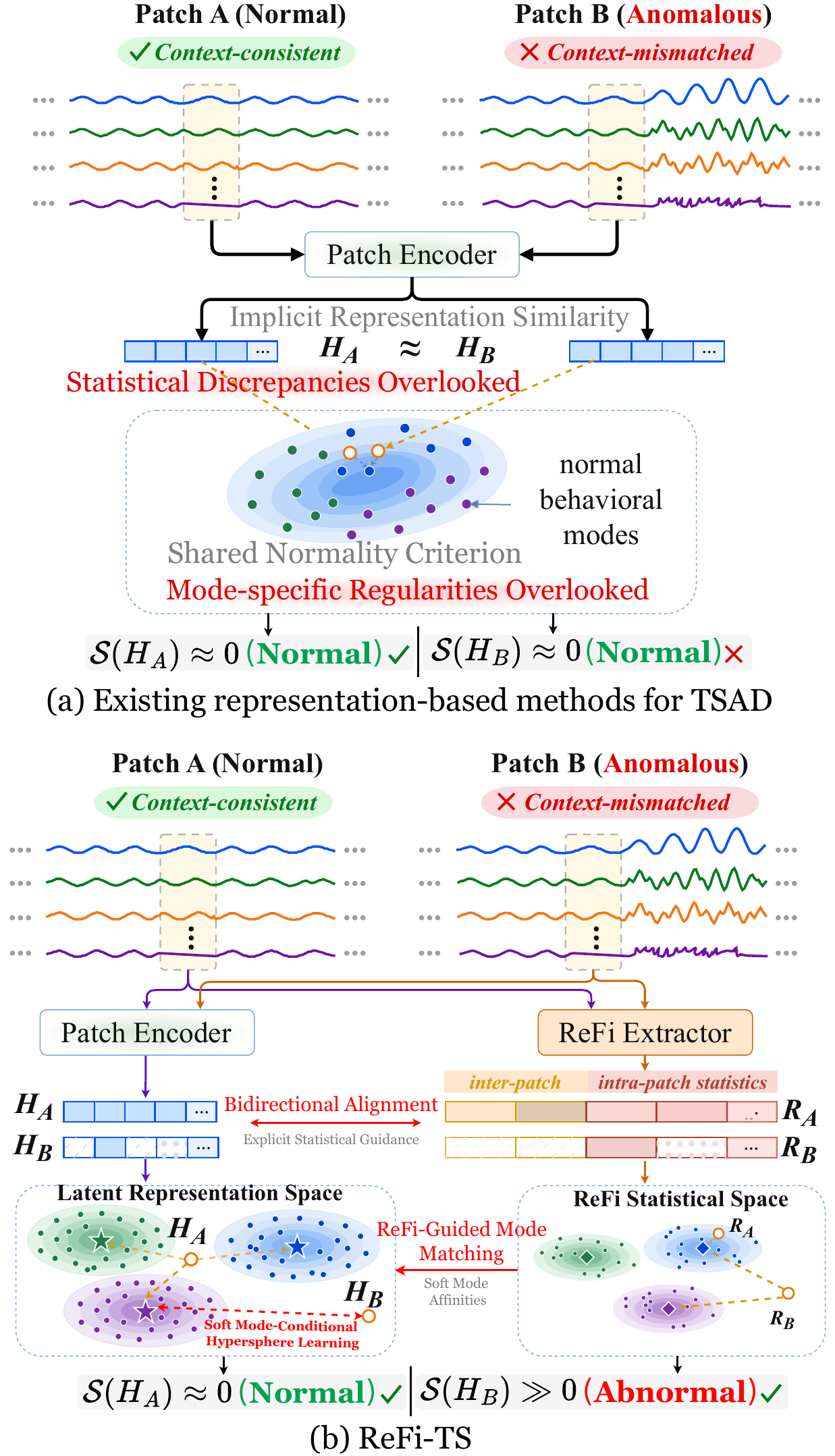}
\caption{Comparison of existing representation-based methods 
and ReFi-TS. 
(a)  Existing methods overlook explicit 
statistical discrepancies and mode-specific normality 
through implicit representation learning.
(b) ReFi-TS incorporates explicit statistical relationships 
and intrinsic temporal structures into deep representation 
learning, enabling mode-conditional normality modeling 
and complementary anomaly detection.}
\label{fig:motivation}
\end{figure}

\section{Introduction}
Time-series anomaly detection (TSAD) aims to identify observations 
or temporal segments that deviate from normal temporal dynamics 
\cite{xu2022anomalytransformer}.
It has attracted considerable attention across diverse applications, 
including industrial fault diagnosis
\cite{su2019robust,TranAD}, 
financial market surveillance 
\cite{poutre2024deep,safa2024waldata}, 
and healthcare monitoring
\cite{wang2023ecggan,qin2023novel}.
In real-world scenarios, anomalies are typically rare, heterogeneous, and difficult to anticipate, 
making reliable anomaly annotations costly and often unavailable \cite{goswami2023unsupervised,enttsel2024general}.
Consequently, unsupervised TSAD focuses on learning
normal temporal patterns from unlabeled or predominantly
normal data and identifying anomalous observations
through deviations from the learned normality.
 
Existing unsupervised TSAD methods can be broadly categorized into four groups: 
(i) \emph{Statistical} \cite{sabokrou2020deep,goyal2020drocc,liu2008isolation} and 
\emph{density estimation} \cite{yairi2017data,zong2018deep} methods identify anomalies 
by modeling deviations from normal data distributions; 
(ii) \emph{Reconstruction and forecasting} methods detect anomalies through reconstruction errors or prediction residuals
\cite{park2018multimodal,su2019robust,zhou2019beatgan,song2024memto,li2021multivariate,lai2024nominality,
hundman2018detecting,luo2024moderntcn,liu2023itransformer,zhang2023crossformer}; 
(iii) \emph{Clustering and distance-based} methods measure deviations from normal clusters 
or neighboring samples in feature spaces \cite{ruff2018deep,shen2020timeseries}; 
and (iv) \emph{Association and discrepancy learning} methods capture anomalies through inconsistencies 
in temporal dependencies, 
inter-variable associations, or cross-view representations\cite{xu2022anomalytransformer,yang2023dcdetector}.
Recently, patch-based representation learning has received 
increasing attention in TSAD for its ability to capture 
subsequence-level temporal patterns.
These methods partition time series into patches and learn 
discriminative temporal representations through specialized 
encoding architectures and representation learning objectives.
Anomalies are subsequently identified by measuring deviations 
from learned normal patterns in the latent space.
PatchAD \cite{zhong2025patchad} employs a multi-scale patch-based 
MLP-Mixer with contrastive learning, while PaAno 
\cite{park2026paano} combines lightweight CNN-based patch encoding 
with triplet and pretext learning objectives to enhance 
representation discriminability.
PaAno+ \cite{zhu2026paano+} further incorporates multi-scale 
encoding and cross-variable attention to enrich patch representations.

Despite their competitive performance, existing representation-based 
methods predominantly rely on implicit feature similarities 
without explicitly modeling the underlying statistical semantics.
As illustrated in Fig.~\ref{fig:motivation}(a), similarity in 
the latent space does not necessarily imply consistency in 
contextual statistical relationships or intrinsic temporal structures.
Specifically, variations in temporal levels and dispersion relative 
to surrounding reference segments, as well as differences in 
trend, distributional shape, and autocorrelation, are not necessarily preserved in the learned representations.
Consequently, latent feature similarity alone can obscure 
statistically meaningful deviations indicative of anomalies.
A further challenge arises from the heterogeneity of normal 
temporal behaviors.
Normal time series often exhibit multiple behavioral modes
with distinct statistical characteristics and
temporal structures.
A shared global normality criterion tends to obscure mode-specific 
regularities, while instance-level discrimination alone does not 
explicitly capture statistical affinities between patches 
and normal modes.
Moreover, mode assignment alone is insufficient to ensure conformity 
to the corresponding mode-specific representation distribution.

To address these limitations, we propose  \textbf{ReFi-TS}, 
a temporal \textbf{\underline{Re}}lational \textbf{\underline{Fi}}ngerprint-guided normality learning method for 
\textbf{\underline{T}}ime \textbf{\underline{S}}eries anomaly detection,
as illustrated in Fig.~\ref{fig:motivation}(b). 
ReFi-TS constructs a five-dimensional temporal relational 
fingerprint (ReFi) to explicitly quantify contextual statistical 
relationships and intrinsic temporal structures of each patch.
Level deviation and log variance ratio quantify contextual statistical variations relative 
to non-overlapping temporal reference segments, 
whereas normalized trend, excess kurtosis, and lag-1 autocorrelation encode intra-patch
temporal structures.
A gated ReFi encoder separately encodes and adaptively
integrates contextual and structural descriptors,
while a bidirectional cross-view alignment mechanism
establishes correspondence between ReFi-derived
representations and deep patch representations.
Furthermore, to characterize heterogeneous normal behaviors, 
we propose a ReFi-guided normality learning mechanism. 
ReFi prototype-based assignments capture patch-to-mode associations, 
which are transferred to deep representations through a ReFi-guided Comparator. 
Soft mode-conditional hypersphere learning further constrains mode-specific 
representation distributions, while hard-negative contrastive learning enhances instance-level discrimination.
Finally, we introduce a complementary anomaly scoring mechanism 
that combines deep representation discrepancy, global ReFi rarity, 
and local ReFi disaffinity through a two-stage fusion strategy, 
jointly assessing deviations in deep representations, global 
statistical patterns, and local temporal contexts.

The main contributions of this work are summarized as follows:
\begin{itemize}
\item  We introduce a temporal relational fingerprint to explicitly characterize contextual statistical relationships 
and intrinsic temporal structures within time-series patches, 
providing complementary statistical semantics beyond implicit deep representations.
\item We propose a ReFi-guided normality learning method that transfers statistical guidance to deep patch representations,
with heterogeneous normal behaviors further characterized through ReFi prototype-guided mode matching 
and soft mode-conditional hypersphere learning.
\item We develop a complementary anomaly scoring mechanism that integrates deep representation discrepancy,
global ReFi rarity, and local ReFi disaffinity.
Experiments on TSB-AD-U and TSB-AD-M demonstrate
consistent improvements over state-of-the-art baselines.
\end{itemize}

\section{Related work}
\paragraph{\textbf{Unsupervised Time Series Anomaly Detection.}}
Unsupervised time series anomaly detection establishes anomaly 
detection criteria by characterizing the underlying regularities 
of normal temporal behaviors without anomaly annotations.
Classical anomaly detection methods identify anomalies by modeling 
statistical distributions \cite{li2020copod,yairi2017data}, 
estimating local densities \cite{breunig2000lof}, 
measuring inter-instance distances 
\cite{ramaswamy2000efficient,paparrizos2015k}, 
or isolating anomalous observations \cite{liu2008isolation}.
With the advances in deep learning, various learning paradigms 
have been explored to characterize temporal normality.
Reconstruction-based 
\cite{audibert2020usad,su2019robust,li2021multivariate,
park2018multimodal,zhou2019beatgan,lai2024nominality}
and forecasting-based methods
\cite{hundman2018detecting,liu2023itransformer,
zhang2023crossformer,luo2024moderntcn}
quantify anomalous deviations through reconstruction 
or prediction residuals.
Several methods further exploit the distributional characteristics 
of normal representations to facilitate anomaly detection.
THOC \cite{shen2020timeseries} models heterogeneous normal 
patterns through hierarchical one-class learning, while 
MEMTO \cite{song2023memto} employs a learnable memory module 
to capture representative normal patterns for sequence reconstruction.
Dissimilarity-based methods establish anomaly detection criteria 
by exploiting discrepancies in temporal associations 
\cite{xu2022anomalytransformer} or cross-view representations 
\cite{yang2023dcdetector}.
CoAD \cite{tang2026bridging} jointly exploits classification 
and reconstruction objectives for anomaly detection, 
while STREAM-VAE \cite{ozer2026stream} employs dual-path 
variational encoding to characterize slow and fast temporal dynamics.
Recent studies also investigate the transferability of pre-trained 
temporal representations for anomaly detection.
TSPulse \cite{ekambaram2026tspulse} learns disentangled temporal, 
spectral, and semantic representations, while TimeRCD 
\cite{lan2026timercd} exploits relative context discrepancy 
learning for zero-shot anomaly detection.

\begin{figure*}[t]
\centering
\includegraphics[width=0.95\textwidth]{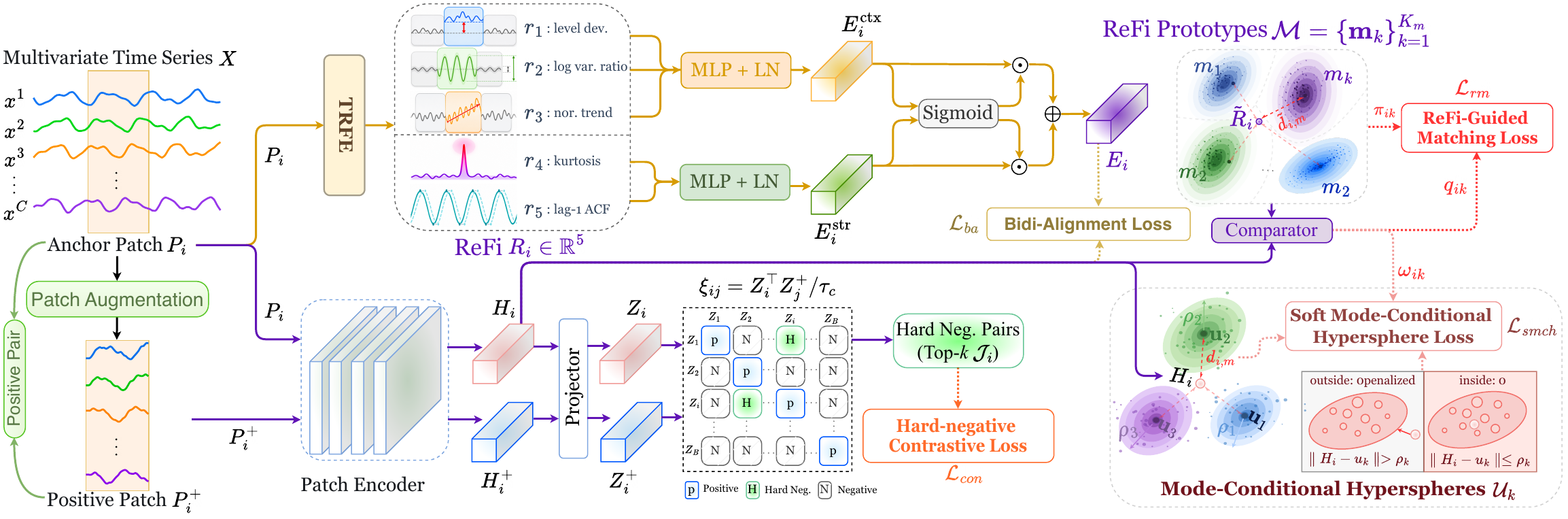}
\vspace{-0.3cm}
\caption{Overview of ReFi-TS.
For each time-series patch, a five-dimensional ReFi 
is extracted to explicitly characterize contextual statistical relationships and intrinsic temporal structures.
A gated ReFi encoder and a patch encoder learn complementary statistical and deep representations, respectively, 
with bidirectional cross-view alignment.
To model heterogeneous normality, ReFi-based prototypes provide soft normal-mode assignments,
 which are transferred to deep representations through a ReFi-guided Comparator.
The resulting mode affinities further guide soft mode-conditional hypersphere learning, 
imposing mode-specific geometric constraints on deep representations.}
\label{overview}
\end{figure*}

Despite their competitive performance,  conventional statistical methods establish detection criteria 
in predefined feature spaces, whereas deep learning-based methods 
predominantly derive anomaly evidence from learned temporal patterns 
and representations.
Explicit temporal statistical relationships, however, remain 
underexplored as guidance for deep normality learning.
In contrast, ReFi-TS leverages temporal relational fingerprints 
to guide the learning of discriminative normal representations, 
while deriving complementary anomaly evidence from deep 
representation and explicit statistical spaces.

\textit{\textbf{Representation Learning for TSAD.}}
Time series representation learning seeks to encode temporal 
dynamics and underlying dependencies into informative 
representations for diverse downstream tasks
\cite{franceschi2019unsupervised,tonekaboni2021unsupervised,
EldeleEmadeldeen2021TRLv,yue2022ts2vec,woo2022cost,
zhang2022self,kara2026adats}.
Among representative methods, TS2Vec~\cite{yue2022ts2vec} 
employs hierarchical contrastive learning to capture contextual 
representations at multiple temporal scales.
TF-C~\cite{zhang2022self} exploits time-frequency consistency 
to learn transferable temporal representations, 
while AdaTS \cite{kara2026adats} introduces dynamic contrastive 
learning to preserve relative instance similarities and 
accommodate non-stationary temporal dynamics.

For time series anomaly detection,  representation learning 
establishes detection criteria by characterizing normality 
and quantifying anomalous deviations in the latent space~\cite{shen2020timeseries,carmona2021neural,
yang2023dcdetector,tang2024perturbation,xu2024calibrated,DIVAD,
darban2025dacad,yin2026scatterad,zhong2025patchad,park2026paano,zhu2026paano+}.
CARLA~\cite{darban2025carla} and 
PCRTA~\cite{tang2024perturbation} enhance representation 
discriminability through contrastive learning with anomaly 
injection and perturbation-guided sample selection, respectively.
DCdetector~\cite{yang2023dcdetector} and ScatterAD~\cite{yin2026scatterad} 
derive anomaly criteria from representation discrepancies across complementary views.
Recent studies further investigate patch-level representation 
learning for fine-grained anomaly detection.
PatchAD~\cite{zhong2025patchad} models temporal patterns 
through multi-scale patch interactions and contrastive learning.
PaAno \cite{park2026paano} constructs a discriminative 
patch-level representation space using a lightweight encoder 
with triplet and pretext learning objectives, 
while PaAno+ \cite{zhu2026paano+} further incorporates 
multi-scale encoding and cross-variable attention.

The aforementioned representation-based methods predominantly rely on 
implicit feature similarities or latent distribution constraints 
to characterize temporal normality.
However, proximity in the latent space does not necessarily 
imply consistency in contextual statistical relationships 
or intrinsic temporal structures, potentially obscuring 
anomaly-relevant discrepancies.
Moreover, instance-level discrimination alone provides 
insufficient guidance for characterizing statistical affinities 
to heterogeneous normal modes and their mode-specific 
representation distributions.
In contrast, ReFi-TS exploits explicit temporal relational 
fingerprints to guide deep patch representation learning 
through bidirectional cross-view alignment.
ReFi prototype-guided mode matching, coupled with soft 
mode-conditional hypersphere learning, further characterizes 
normal-mode affinities and within-mode conformity.

\section{Preliminaries}
\label{sec:preliminaries}
\paragraph{\textbf{Problem Formulation.}}
Let $X=(\mathbf{x}_0,\ldots,\mathbf{x}_{T-1})^{\top}
\in\mathbb{R}^{T\times C}$ denote a time series,
where $T$ is the sequence length and 
$\mathbf{x}_t\in\mathbb{R}^{C}$ represents the observation 
of $C$ variables at time step $t$.
The cases $C=1$ and $C>1$ correspond to univariate 
and multivariate time series, respectively.
Given a normal training time series
$X^{\mathrm{tr}}\in\mathbb{R}^{T\times C}$, 
the objective of unsupervised time series anomaly detection is to learn 
an anomaly detector $f_{\Theta}$ that characterizes 
the underlying normal behaviors from $X^{\mathrm{tr}}$ 
and identifies anomalous observations in the test time series 
$X^{\mathrm{te}}\in\mathbb{R}^{T\times C}$.
Formally, the anomaly scoring function is defined as:
\begin{equation}
    \mathcal{S}^{\mathrm{te}}
    = f_{\Theta}(X^{\mathrm{te}})
    \in\mathbb{R}^{T},
    \label{eq:task_formulation}
\end{equation}
where $\Theta$ denotes the model parameters and reference 
statistics estimated from $X^{\mathrm{tr}}$, and 
$\mathcal{S}_t^{\mathrm{te}}$ measures the degree to which the observation  $x_t$
at time step $t$ deviates from learned normal behaviors.
The corresponding ground-truth labels 
$\mathbf{Y}^{\mathrm{te}}\in\{0,1\}^{T}$ of $X^{te}$
are used exclusively for evaluation, with 
$y_t^{\mathrm{te}}=1$ indicating an anomalous observation.

\textbf{\textit{Temporal Patching.}}
To characterize fine-grained temporal patterns, we partition 
a time series $X$ into overlapping patches using a sliding 
window of length $W$ with unit stride.
The resulting patch set is denoted by
$\mathcal{P}=\{P_i\}_{i=0}^{N-1}$, where $N=T-W+1$ is the number of patches.
Each patch is formulated as:
\begin{equation}
    P_i=[\mathbf{x}_i,\ldots,\mathbf{x}_{i+W-1}]
    \in\mathbb{R}^{W\times C},
    \quad 0\leq i<N.
    \label{eq:patch}
\end{equation}
Accordingly, the training and test time series yield patch sets
$\mathcal{P}^{\mathrm{tr}}$ and $\mathcal{P}^{\mathrm{te}}$,
respectively.
The detector learns discriminative temporal representations 
from training patches, while anomaly scores of test patches 
are aggregated over their overlapping temporal positions 
to obtain point-wise anomaly scores.

\section{Methodology}
\label{sec:methodology}
\subsection{Overview}
\label{sec:overview}
ReFi-TS establishes a statistically informed normality learning 
framework by incorporating explicit temporal statistical relationships 
into deep patch representation learning.
As illustrated in Fig.~\ref{overview}, ReFi-TS first constructs 
a temporal relational fingerprint (ReFi) to explicitly characterize 
contextual statistical relationships and intrinsic temporal structures 
of each patch (Sec.~\ref{sec:refi_extraction}).
Bidirectional cross-view alignment establishes correspondence 
between ReFi embeddings and contrastively learned patch representations, 
thereby introducing explicit statistical guidance into deep representation 
learning (Sec.~\ref{sec:refi_alignment}).
To characterize heterogeneous normal behaviors, we develop ReFi
prototype-guided mode matching coupled with soft mode-conditional
hypersphere learning to jointly model normal-mode affinities 
and within-mode conformity (Sec.~\ref{sec:normality_modeling}).
During inference, deep representation discrepancy, global ReFi rarity, 
and local ReFi disaffinity are jointly exploited to derive 
complementary anomaly scores (Sec.~\ref{sec:training_inference}).

\subsection{Temporal Relational Fingerprint Extraction}
\label{sec:refi_extraction}
The same temporal pattern may have different anomaly implications depending on
its surroundings. Consider two identical patches $P_i=P_j$ embedded in distinct
operating contexts. A deterministic patch encoder in evaluation mode necessarily
produces $f_\theta(P_i)=f_\theta(P_j)$, even when the two patches have different
relationships with their respective neighborhoods.
Conversely, two patches with identical means and variances may exhibit distinct
internal dynamics; reversing the temporal order, for example, reverses the sign
of a nonzero linear trend without changing either statistic.
These observations motivate a descriptor that explicitly captures both
inter-patch statistical relationships and intra-patch structure.
Inspired by the relational fingerprint perspective in graph anomaly detection
\cite{REFI-GAD}, we construct a five-dimensional temporal relational
fingerprint, ReFi, for each patch. Unlike its graph counterpart, ReFi is derived
from ordered observations and temporal references, and serves as statistical
guidance for deep learning rather than a replacement for the original input.

\emph{\textbf{a) Contextual Statistical Relationships.}}
The significance of a level shift or a change in dispersion depends on the
local operating regime. To establish a contextual reference, we first derive
a representative signal for each patch $P_i$ by channel averaging:
\begin{equation}
    m_{i,\tau}=\frac{1}{C}\sum_{c=1}^{C}P_i[\tau,c],
    \quad 0\leq\tau<W.
    \label{eq:refi_channel_mean}
\end{equation}
The corresponding patch-level mean and population variance are computed as:
\begin{equation}
    \mu_i =\frac{1}{W}\sum_{\tau=0}^{W-1}m_{i,\tau}, \quad 
    v_i=\frac{1}{W}\sum_{\tau=0}^{W-1}(m_{i,\tau}-\mu_i)^2.
\label{eq:refi_patch_statistics}
\end{equation}
This yields the original signal for univariate series and an aggregate signal
for multivariate series. Channel-specific patterns remain accessible to the deep
encoder, which processes the original multichannel patch.

Under unit-stride patching, consecutive patches share $W-1$ observations.
Their statistical agreement can therefore be induced by sample overlap rather
than genuine contextual consistency. 
We therefore select temporal references
at integer multiples of $W$. The non-overlapping reference indices for patch $P_i$ are defined as:
\begin{equation}
    \mathcal{N}_i=\{i+qW\mid q\in\mathbb{Z},\ 1\leq |q|\leq h,0\leq i+qW<N\},
\label{eq:refi_reference_set}
\end{equation}
where $h$ denotes the maximum number of references on either side of $P_i$.
The corresponding reference mean and average within-reference variance are computed as:
\begin{equation}
    \bar{\mu}_i=\frac{1}{\left\lvert \mathcal{N}_i\right\rvert }\sum_{j\in\mathcal{N}_i}\mu_j, \quad
    \bar{v}_i=\frac{1}{\left\lvert \mathcal{N}_i\right\rvert }\sum_{j\in\mathcal{N}_i}v_j,
\label{eq:refi_reference_statistics}
\end{equation}
where $\bar v_i$ is the average within-reference variance, excluding variation
between reference means.
If no valid reference exists, both reference statistics are set to zero.
We derive two complementary descriptors from these statistics.

\textbf{\textit{\italiccircled[black]{1} Dim 1: Relative Level Deviation.}}
To distinguish a level shift from contextual fluctuations, 
we quantify the level deviation by normalizing the signed difference between the patch and reference means 
with the reference fluctuation scale, defined as:
\begin{equation}
    r_{i,1}=\frac{\mu_i-\bar{\mu}_i}{\sqrt{\bar{v}_i+\varepsilon}},
    \label{eq:refi_level}
\end{equation}
where $\varepsilon>0$ is a numerical stabilization constant.
The magnitude quantifies contextual level deviation, whereas the sign
distinguishes upward from downward shifts.

\textbf{\textit{\italiccircled[black]{2} Dim 2: Log Variance Ratio.}}
A patch may exhibit atypical variability despite maintaining
a stable operating level. We quantify this deviation using
the log variance ratio:
\begin{equation}
    r_{i,2}=\log\frac{v_i+\varepsilon}{\bar{v}_i+\varepsilon},
    \label{eq:refi_variability}
\end{equation}
where positive and negative values indicate increased and decreased
variability relative to the reference, respectively.
The two dimensions thus distinguish relative location shifts from changes in
dispersion without relying on absolute operating levels.

\emph{\textbf{b) Intrinsic Temporal Structures.}}
The contextual descriptors characterize patch-level location and
dispersion but are invariant to permutations of observations
within a patch. We therefore introduce three intrinsic descriptors
to capture temporal trend, amplitude-distribution shape, and
short-range dependence.
 
\textbf{\textit{\italiccircled[black]{3} Dim 3: Normalized Trend.}}
To capture directional evolution within a patch, we estimate
its least-squares slope. Let $\bar\tau=(W-1)/2$ denote the
mean time index and $a_{i,\tau}=m_{i,\tau}-\mu_i$ the
centered observation. The slope is computed as:
\begin{equation}
    \beta_i=\frac{\sum_{\tau=0}^{W-1}(\tau-\bar\tau)a_{i,\tau}}
    {\sum_{\tau=0}^{W-1}(\tau-\bar\tau)^2+\varepsilon}.
    \label{eq:refi_slope}
\end{equation} 
We normalize the slope by the within-patch fluctuation scale
to obtain:
\begin{equation}
    r_{i,3}=\frac{W\beta_i}{\sqrt{v_i+\varepsilon}}.
    \label{eq:refi_trend}
\end{equation}
This descriptor captures the direction and relative magnitude
of linear change independently of the absolute fluctuation scale.

\textbf{\textit{\italiccircled[black]{4} Dim 4: Excess Kurtosis.}}
To characterize distributional deviations beyond variance,
we compute the regularized excess kurtosis:
\begin{equation}
    r_{i,4}=\frac{W^{-1}\sum_{\tau=0}^{W-1}a_{i,\tau}^{4}}
    {v_i^2+\varepsilon}-3,
    \label{eq:refi_kurtosis}
\end{equation}
where the subtraction of $3$ yields zero excess kurtosis
for a Gaussian distribution. Larger positive values indicate
more pronounced extreme deviations, providing a complementary
measure of amplitude-distribution characteristics.

\textbf{\textit{\italiccircled[black]{5} Dim 5: Lag-1 Autocorrelation.}}
To capture temporal dependencies not reflected in marginal
amplitude statistics, we quantify adjacent-sample dependence as:
\begin{equation}
    r_{i,5}=\frac{\sum_{\tau=0}^{W-2}a_{i,\tau}a_{i,\tau+1}}
    {\sum_{\tau=0}^{W-1}a_{i,\tau}^{2}+\varepsilon},
    \label{eq:refi_autocorrelation}
\end{equation}
where positive and negative values indicate predominantly persistent and
alternating deviations from the patch mean, respectively. 
Unlike the marginal moments, this descriptor explicitly captures
local temporal ordering.

The five descriptors are concatenated to form the temporal relational fingerprint:
\begin{equation}
    R_i=[r_{i,1},r_{i,2},r_{i,3},r_{i,4},r_{i,5}]^{\top} 
    \in\mathbb{R}^{5}.
    \label{eq:refi_vector}
\end{equation}
The first two descriptors describe how a patch differs from its context, and the remaining three characterize how its
observations are distributed and organized. We treat their joint configuration, rather than any individual extreme
value, as the statistical signature of a temporal pattern. 
The fingerprints are standardized using component-wise statistics estimated from normal training data:
\begin{equation}
    \tilde R_i=(R_i-\boldsymbol\mu_R)\oslash
    \max(\boldsymbol\sigma_R,\varepsilon_R),
    \label{eq:refi_standardization}
\end{equation}
where $\boldsymbol\mu_R$ and $\boldsymbol\sigma_R\in\mathbb{R}^5$ denote the
component-wise training mean and sample standard deviation, respectively.
Both $\oslash$ and $\max$ act element-wise, and $\varepsilon_R>0$ prevents
division by zero.
The same statistics are retained at inference.

\subsection{ReFi-Guided Patch Representation Learning}
\label{sec:refi_alignment}
ReFi provides a compact and interpretable characterization of contextual statistical relationships 
and intrinsic temporal structures, 
complementing deep representations that capture richer multivariate temporal dynamics. 
While ReFi focuses on selected statistical properties, 
deep patch representations do not necessarily preserve their underlying statistical semantics. 
To exploit their complementary strengths, we learn both views independently 
and establish correspondence at the representation level, 
rather than directly concatenating ReFi with the encoder input. 
By integrating instance-level contrastive learning with bidirectional cross-view alignment, 
our method encourages discriminative temporal representations while explicitly 
incorporating statistical semantics into the learning process.

\textbf{\textit{a) Deep Patch Representation and Gated ReFi Encoding.}}
For each patch $P_i$, we obtain its deep representation through a CNN encoder:
\begin{equation}
    H_i=f_{\theta}(P_i)=\operatorname{GAP}(\operatorname{CNN}_{\theta}(P_i)),
    \label{eq:temporal_embedding}
\end{equation}
where $H_i\in\mathbb{R}^d$, $\operatorname{CNN}_{\theta}$ comprises four one-dimensional convolutional
blocks with batch normalization and ReLU, and $\operatorname{GAP}$ denotes global
average pooling along the temporal dimension.

Contextual deviations and intrinsic structural changes capture
complementary aspects of temporal behavior, whose relative
importance may vary across patches. To preserve their distinct
semantics, we independently encode the two groups of ReFi descriptors:
\begin{equation}
\begin{aligned}
    E_i^{\mathrm{ctx}}&=\operatorname{LN}(g_{\mathrm{ctx}}(\tilde R_{i,1:2})) \in \mathbb{R}^d,\\
    E_i^{\mathrm{str}}&=\operatorname{LN}(g_{\mathrm{str}}(\tilde R_{i,3:5})) \in \mathbb{R}^d,
\end{aligned}
\label{eq:fingerprint_subembeddings}
\end{equation}
where $g_{\mathrm{ctx}}(\cdot)$ and $g_{\mathrm{str}}(\cdot)$ are separate two-layer MLPs,
and $\operatorname{LN}$ denotes layer normalization.
To adaptively balance their contributions, we introduce
a feature-wise gating mechanism:
\begin{equation}
\begin{aligned}
    G_i&=\operatorname{sigmoid}\!\left(A_g[E_i^{\mathrm{ctx}}\Vert E_i^{\mathrm{str}}]+b_g\right),\\
    E_i&=G_i\odot E_i^{\mathrm{ctx}}+(\mathbf{1}-G_i)\odot E_i^{\mathrm{str}},
\end{aligned}
\label{eq:fingerprint_gated_fusion}
\end{equation}
where $A_g\in\mathbb{R}^{d\times 2d}$ and $b_g\in\mathbb{R}^d$ are learnable,
$\Vert$ denotes concatenation, and $\odot$ denotes element-wise multiplication.
The resulting fused ReFi representation $E_i\in\mathbb{R}^d$
is dimensionally aligned with $H_i$, facilitating subsequent
cross-view correspondence learning.
 
\textbf{\textit{b) Hard Negative Contrastive Learning.}}
To learn discriminative patch representations while preserving
robustness to minor perturbations, we construct a positive view
$P_i^{+}$ for each anchor patch $P_i$ by introducing Gaussian
noise and a small channel-wise offset, 
with both perturbations scaled by the within-patch standard deviation.
The two views are processed by a shared encoder and projected
into a normalized contrastive space:
\begin{equation}
    Z_i = \operatorname{norm}\left(g_{\mathrm{pr}}(H_i)\right) \in\mathbb{R}^{d_p},
    \quad
    Z_i^{+} = \operatorname{norm}\left(g_{\mathrm{pr}}(H_i^{+})\right) \in\mathbb{R}^{d_p},
    \label{eq:contrastive_projection}
\end{equation}
where $H_i=f_{\theta}(P_i)$ and
$H_i^{+}=f_{\theta}(P_i^{+})$,
$g_{\mathrm{pr}}$ denotes a two-layer projector,
and $\operatorname{norm}$ performs $\ell_2$ normalization.

To emphasize challenging negatives, we measure the
temperature-scaled similarity between the anchor and
augmented views of other patches as
$\xi_{ij}=Z_i^{\top}Z_j^{+}/\tau_c$, where $\tau_c>0$
is the temperature parameter.
We then select the $k_h$ most similar nonmatching views
within a mini-batch $\mathcal{B}$ of size $B$:
\begin{equation}
  \mathcal{J}_i =
    \operatorname{TopK}_{j\in\mathcal{B}\setminus\{i\}}
    \left(\xi_{ij},k_h\right).
    \label{eq:hard_negative_selection}
\end{equation}
Let $\mathcal{C}_i=\{i\}\cup\mathcal{J}_i$ denote the
candidate set and $L_i=|\mathcal{J}_i|$ its number of
hard negatives. We formulate the hard-negative contrastive objective
with label smoothing as:
\begin{equation}
    \mathcal{L}_{\mathrm{con}}
     = -\frac{1}{B}
    \sum_{i\in\mathcal{B}}
    \sum_{j\in\mathcal{C}_i}
    t_{ij}\log p_{ij}^{\mathrm{con}} 
     = -\frac{1}{B}
    \sum_{i\in\mathcal{B}}
    \sum_{j\in\mathcal{C}_i}
    t_{ij}
    \log
    \frac{\exp(\xi_{ij})}
    {\sum_{\ell\in\mathcal{C}_i}\exp(\xi_{i\ell})},
\label{eq:contrastive_loss}
\end{equation}
where $t_{ij}=(1-\delta)\mathbb{I}[j=i]
    +\frac{\delta}{L_i}\mathbb{I}[j\in\mathcal{J}_i]$,
$\delta$ denotes the label-smoothing coefficient, and
$\mathbb{I}[\cdot]$ is the indicator function.
When $\delta=0$, the objective reduces to InfoNCE \cite{oord2018representation}
over the positive pair and selected hard negatives.
Since training involves only normal patches, the selected
hard negatives correspond to other normal instances rather
than anomalous observations. This objective promotes
instance-level discrimination but does not explicitly
preserve statistical semantics or model heterogeneous
normal modes, motivating the subsequent ReFi-guided
representation learning.

\textbf{\textit{c) Bidirectional Cross-View Alignment.}}
Although contrastive learning promotes instance-level
discriminability, it does not explicitly incorporate
the statistical semantics encoded by ReFi.
We therefore establish bidirectional correspondence
between deep patch representations and ReFi embeddings,
enabling statistical semantics to guide deep representation
learning without enforcing feature-wise equivalence.
Specifically, we define a cross-view similarity matrix
within each mini-batch $\mathcal{B}$ as:
\begin{equation}
    A_{ij} =
    \frac{
        \operatorname{norm}(H_i)^{\top}
        \operatorname{norm}(E_j)
    }{\tau_s},
    \quad i,j\in\mathcal{B},
    \label{eq:sample_matching_score}
\end{equation}
where $E_j$ denotes the ReFi embedding of patch $j$,
$\operatorname{norm}$ performs $\ell_2$ normalization,
and $\tau_s>0$ is a temperature parameter.
The diagonal entries correspond to matched cross-view pairs,
while off-diagonal entries represent nonmatching instances.

To enforce correspondence in both directions, we minimize a symmetric cross-view alignment objective:
\begin{equation}
    \mathcal{L}_{\mathrm{ba}}
    =-\frac{1}{2B}\sum_{i\in\mathcal{B}}\Bigg[
     \log
    \frac{\exp(A_{ii})}
    {\sum_{j\in\mathcal{B}}\exp(A_{ij})}
    + \log
    \frac{\exp(A_{ii})}
    {\sum_{j\in\mathcal{B}}\exp(A_{ji})}
    \Bigg].
\label{eq:sample_alignment_loss}
\end{equation}
The two terms respectively enforce deep-to-ReFi and
ReFi-to-deep correspondence, encouraging matched
representations to exhibit consistent cross-view
associations. Unlike direct feature matching, this
objective preserves the flexibility of both representation
spaces while promoting statistically informed
deep representations.

\subsection{ReFi-Guided Normality Modeling}
\label{sec:normality_modeling}
Instance discrimination promotes the separability of individual
patches but does not explicitly capture the underlying structure
of normal behavior.
In practice, normal patches may share similar statistical
characteristics while exhibiting substantial variations
in their temporal dynamics.
We therefore extend ReFi guidance to normality modeling through
two complementary properties.
\emph{Mode affinity} characterizes the statistical
association of each patch with different normal modes
and guides deep representations to preserve these
associations.
\emph{Within-mode conformity} further constrains
deep representations to reside within the characteristic
geometric regions of their associated normal modes.

\textbf{\textit{a) ReFi-Guided Mode Matching.}}
To establish stable statistical references for normal
behaviors, we apply MiniBatch K-means to the standardized
ReFi descriptors of normal training patches, avoiding
direct clustering in the evolving deep representation space.
The resulting prototype set is denoted by
$\mathcal{M}=\{\mathbf{m}_k\}_{k=1}^{K_m}$,
where $\mathbf{m}_k\in\mathbb{R}^{5}$.
For each patch, we quantify its ReFi-based affinity
to the normal-mode prototypes through distance-based
soft assignments:
\begin{equation}
    \pi_{ik}
    =
    \frac{
        \exp(-\|\tilde R_i-\mathbf{m}_k\|_2^2/\tau_a)
    }{
        \sum_{\ell=1}^{K}
        \exp(-\|\tilde R_i-\mathbf{m}_{\ell}\|_2^2/\tau_a)
    },
    \label{eq:fingerprint_mode_assignment}
\end{equation}
where $\tilde R_i$ denotes the standardized ReFi descriptor
of patch $i$ and $\tau_a>0$ controls the assignment concentration.
The resulting distribution
$\pi_i=(\pi_{i1},\ldots,\pi_{iK_m})^\top$
characterizes the statistical mode affinity of patch $i$
in the ReFi space.

Since ReFi prototypes and deep representations reside
in feature spaces with distinct dimensionalities,
direct feature-wise matching is unsuitable.
Therefore, we introduce a shared ReFi-guided
\emph{Comparator} to estimate the compatibility
between each deep representation $H_i$ and
statistical prototype $\mathbf{m}_k$:
\begin{equation}
\begin{aligned}
    e_{ik}
    &=\operatorname{Comparator}
    ([H_i\Vert\mathbf{m}_k]),\\
    q_{ik}
    &=\frac{\exp(e_{ik}/\tau_p)}
    {\sum_{\ell=1}^{K}\exp(e_{i\ell}/\tau_p)},
\end{aligned}
\label{eq:temporal_mode_assignment}
\end{equation}
where $[\cdot\Vert\cdot]$ denotes concatenation,
the Comparator is a shared two-layer MLP with
a scalar output, and $\tau_p>0$ is the prediction
temperature.
$q_i=(q_{i1},\ldots,q_{iK})^\top$
represents the mode affinities predicted from
the deep representation $H_i$.

We formulate the ReFi-guided mode matching loss, $\mathcal{L}_{\mathrm{rm}}$, 
to supervise the mode-affinity predictions $q_i$ using the ReFi-derived assignments $\pi_i$:
\begin{equation}
    \mathcal{L}_{\mathrm{rm}}
    =
    -\frac{1}{B}
    \sum_{i\in\mathcal{B}}
    \sum_{k=1}^{K_m}
    \pi_{ik}\log q_{ik}.
    \label{eq:mode_alignment_loss}
\end{equation}
Since $\pi_i$ is fixed, minimizing
$\mathcal{L}_{\mathrm{rm}}$ is equivalent,
up to an additive constant, to minimizing
$B^{-1}\sum_{i\in\mathcal{B}}
D_{\mathrm{KL}}(\pi_i\|q_i)$.
This objective transfers ReFi-derived mode supervision
to deep representations, encouraging them to preserve
the statistical associations of normal patches with
different modes without imposing the geometric structure
of the ReFi space.

\textbf{\textit{b) Soft Mode-Conditional Hypersphere Learning.}}
Mode-affinity supervision encourages deep representations
to preserve their ReFi-derived associations with normal
modes, but does not constrain their geometric positions
within the associated modes.
To capture within-mode conformity, we associate each
normal mode with a hypersphere in the deep representation
space:
\begin{equation}
    \mathcal{U}_k
    =
    \left\{
        H\in\mathbb{R}^{d}:
        \|H-\mathbf{u}_k\|_2\leq\rho_k
    \right\},
    \quad k=1,\ldots,K_m,
    \label{eq:mode_hypersphere}
\end{equation}
where $\mathbf{u}_k\in\mathbb{R}^{d}$ and $\rho_k\geq0$
denote the center and radius of the $k$-th hypersphere,
respectively.
Unlike the fixed ReFi prototype $\mathbf{m}_k$,
the hypersphere center $u_k$ adapts to the evolving
geometry of deep representations.

A patch may exhibit affinity to multiple normal
modes, therefore we derive soft mode weights from the Comparator
logits in Eq.~\eqref{eq:temporal_mode_assignment}:
\begin{equation}
    \omega_{ik}
    =
    \operatorname{sg}\left(
    \frac{\exp(e_{ik}/\tau_b)}
    {\sum_{\ell=1}^{K_m}\exp(e_{i\ell}/\tau_b)}
    \right),
    \label{eq:boundary_mode_weights}
\end{equation}
where $\operatorname{sg}(\cdot)$ denotes stop-gradient
and $\tau_b>0$ controls the concentration of the
mode weights.
These weights determine the contribution of each
mode-specific boundary penalty, yielding the soft
mode-conditional hypersphere loss $\mathcal{L}_{\mathrm{smch}}$  as follows:
\begin{equation}
    \mathcal{L}_{\mathrm{smch}}
    =
    \frac{1}{B}
    \sum_{i\in\mathcal{B}}
    \sum_{k=1}^{K_m}
    \omega_{ik}
    \left[
        \|H_i-\mathbf{u}_k\|_2-\rho_k
    \right]_+^2,
    \label{eq:mode_boundary_loss}
\end{equation}
where $[a]_+=\max(a,0)$.
The hinge term penalizes representations only when
they fall outside the corresponding mode-specific
hypersphere, with each penalty weighted by its
predicted mode affinity.
This soft weighting allows a patch to be constrained
by multiple plausible modes, while stop-gradient
prevents the Comparator from reducing the geometric
penalty through mode reassignment.
The hypersphere centers and radii are maintained
separately from gradient-based optimization.

\begin{figure}[t]
    \centering
    \includegraphics[width=\columnwidth]
    {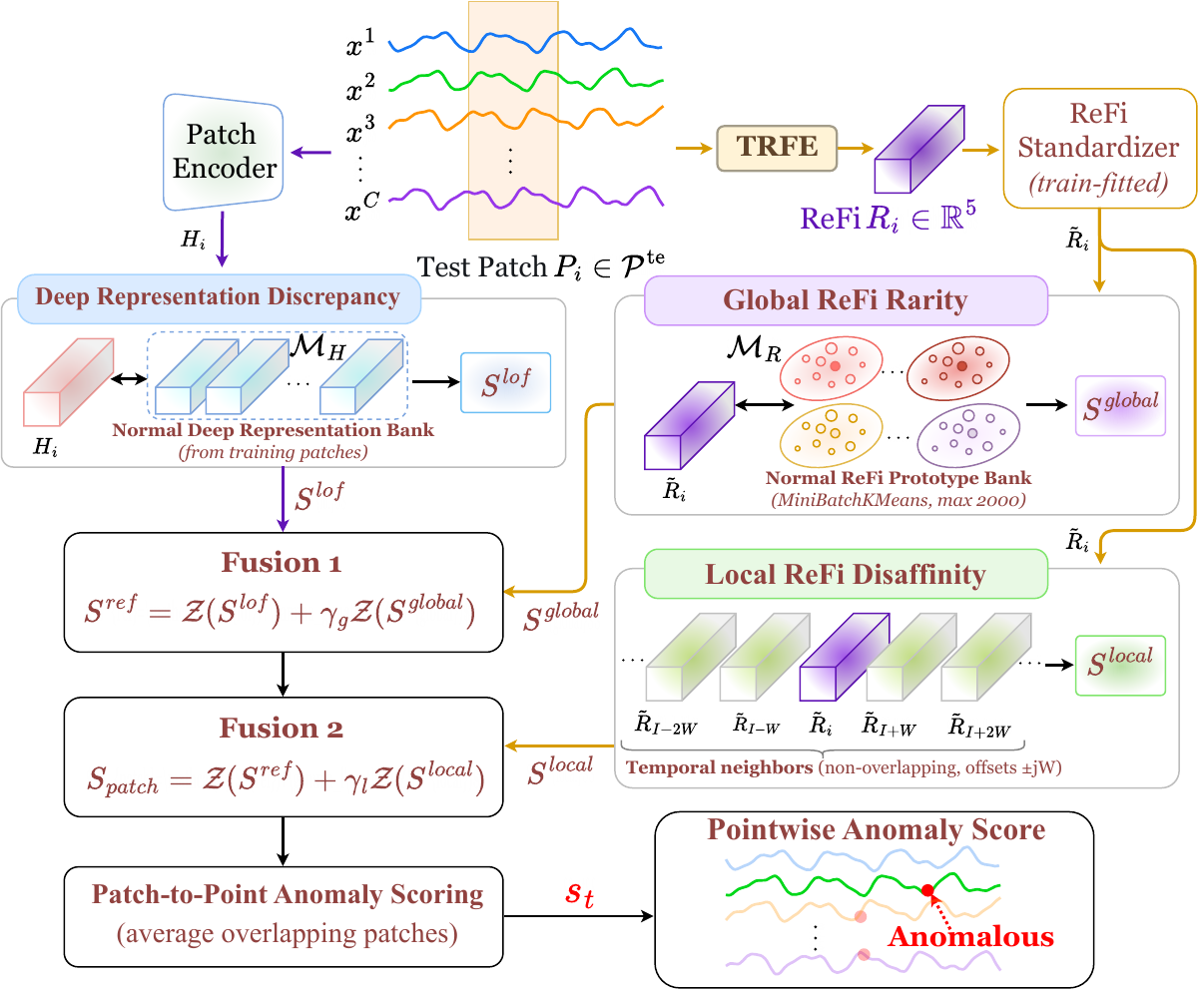}
    \vspace{-0.5cm}
    \caption{Anomaly scoring procedure of ReFi-TS.}
    \label{fig:inference_framework}
\end{figure}

During training, the hypersphere centers and radii are dynamically updated
to track the evolving deep representation space through
exponential moving averages.
Specifically, the Comparator-predicted mode relevance
$q_{ik}$ in Eq.~\eqref{eq:temporal_mode_assignment}
is used to determine the dominant mode of each patch
for updating the hypersphere statistics.
The soft weights $\omega_{ik}$ govern the geometric
penalties, whereas hard assignments derived from
$q_{ik}$ determine the mini-batch subsets for
hypersphere updates.
The corresponding mode index and mini-batch subset are defined as:
\begin{equation}
    \hat{k}_i=\arg\max_{1\leq\ell\leq K_m}q_{i\ell}, 
    \mathcal{B}_k=\{i\in\mathcal{B}:\hat{k}_i=k\},
\label{eq:hypersphere_update_assignment}
\end{equation}
where $\mathcal{B}_k$ contains the patches assigned to
mode $k$ in the current mini-batch.
For each nonempty subset, the mini-batch mean is first
computed to update the hypersphere center.
The radius is subsequently updated using the 95th
percentile of the distances to the updated center:
\begin{equation}
\footnotesize
\begin{aligned}
    \bar{\mathbf H}_k
    &=
    \frac{1}{|\mathcal{B}_k|}
    \sum_{i\in\mathcal{B}_k}H_i,\\
    \mathbf{u}_k
    &\leftarrow
    \alpha\mathbf{u}_k
    +(1-\alpha)\bar{\mathbf H}_k,\\
    \rho_k
    &\leftarrow
    \alpha\rho_k
    +(1-\alpha)
    \operatorname{Quantile}_{0.95}
    \left(
        \left\{
        \sqrt{
        \|H_i-\mathbf{u}_k\|_2^2+\varepsilon
        }:
        i\in\mathcal{B}_k
        \right\}
    \right),
\end{aligned}
\label{eq:hypersphere_update}
\end{equation}
where $\bar{\mathbf H}_k$ denotes the mini-batch
mean representation of mode $k$,
$\alpha\in[0,1)$ is the momentum coefficient,
and $\varepsilon>0$ ensures numerical stability.
Empty subsets retain their previous hypersphere parameters.
The mode-matching objective $\mathcal{L}_{\mathrm{rm}}$
preserves ReFi-derived statistical associations across
normal modes, whereas the geometric regularization
objective $\mathcal{L}_{\mathrm{smch}}$ enforces
within-mode conformity through mode-specific
hypersphere constraints.
Their joint optimization captures heterogeneous
normal modes without imposing a shared global boundary.

\subsection{Overall Objective and Anomaly Scoring}
\label{sec:training_inference}
\textbf{\textit{Overall Learning Objective.}}
The overall training objective of ReFi-TS integrates four complementary objectives:
\begin{equation}
    \mathcal{L}_{\mathrm{total}}
    = \lambda_1\mathcal{L}_{\mathrm{con}}
    + \lambda_2\mathcal{L}_{\mathrm{ba}}
    + \lambda_3\mathcal{L}_{\mathrm{rm}}
    + \lambda_4\mathcal{L}_{\mathrm{smch}},
    \label{eq:joint_training_objective}
\end{equation}
where $\lambda_1$, $\lambda_2$, $\lambda_3$,
and $\lambda_4$ are balancing weights selected via grid search \cite{lu2024caudits,ozyurt2023contrastive}.

\textbf{\textit{Anomaly Criterion.}} 
At inference, ReFi-TS evaluates anomalous behavior
from three complementary perspectives.
Deep representation discrepancy captures relative
local sparsity in the learned feature space,
global ReFi rarity measures statistical deviations
from normal training patterns, and local ReFi
disaffinity quantifies inconsistencies with the
current temporal context.
These scores are subsequently fused to obtain
point-wise anomaly scores, as illustrated in
Fig.~\ref{fig:inference_framework}.

\textbf{\textit{\italiccircled[blue]{1} Deep Representation Discrepancy.}}
Normal patterns may occupy regions with different
local densities in the deep representation space.
Consequently, absolute nearest-neighbor distances
may overestimate the anomaly of legitimate patterns
in sparsely populated regions.
We therefore adopt a Local Outlier Factor (LOF) inspired relative distance
criterion that accounts for the local spacing of
neighboring normal references.

A deep representation reference bank
$\mathcal{M}_H=\{b_j\}_{j=1}^{M_H}$
is constructed from representative normal training
representations, with all embeddings $\ell_2$-normalized
for cosine dissimilarity computation.
For normalized embeddings $a$ and $b$, the cosine
dissimilarity is defined as
$D_H(a,b)=1-a^\top b$.
Let $\mathcal{K}_H(a)$ denote the $k_H$ nearest
entries in $\mathcal{M}_H$ under this dissimilarity.
The local spacing of a reference embedding $b_j$
is estimated by:
\begin{equation}
    \ell_H(b_j)
    =
    \frac{1}{k_H}
    \sum_{b\in\mathcal{K}_H(b_j)}
    D_H(b_j,b),
    \label{eq:reference_local_spacing}
\end{equation}
where $b_j$ itself is excluded from its bank-internal neighborhood.

For a normalized query embedding
$\hat H_i=\operatorname{norm}(H_i)$,
the relative deep-space discrepancy is computed as:
\begin{equation}
    S_i^{\mathrm{lof}}
    =
    \frac{
        \frac{1}{k_H}
        \sum_{b\in\mathcal{K}_H(\hat H_i)}
        D_H(\hat H_i,b)
    }{
        \frac{1}{k_H}
        \sum_{b\in\mathcal{K}_H(\hat H_i)}
        \ell_H(b)+\varepsilon_H
    },
    \label{eq:deep_relative_density}
\end{equation}
where $\varepsilon_H>0$ ensures numerical stability
and $k_H\leq M_H-1$ for $M_H\geq2$.
A larger score indicates greater separation from
neighboring references relative to their local spacing.

\textbf{\textit{\italiccircled[blue]{2} Global ReFi Rarity.}}
Cross-view alignment incorporates statistical guidance
into deep representation learning, but does not guarantee
that all statistical deviations are preserved in the
learned feature space.
We therefore assess global statistical rarity directly
in the ReFi space using distances between each query fingerprint
and its nearest normal references:
\begin{equation}
    S_i^{\mathrm{global}}
    =
    \frac{1}{k_R}
    \sum_{v\in\mathcal{K}_R(\tilde R_i)}
    \|\tilde R_i-v\|_2,
    \label{eq:global_refi_score}
\end{equation}
where $\mathcal{K}_R(\tilde R_i)$ contains
the $k_R$ nearest entries in the statistical reference
bank $\mathcal{M}_R$, with $k_R\leq M_R$. 
A larger score indicates that the statistical
characteristics of the query are less represented
in the normal training patterns.

\textbf{\textit{\italiccircled[blue]{3} Local ReFi Disaffinity.}}
Global statistical familiarity does not necessarily
imply affinity with the current temporal context.
A fingerprint may closely resemble normal training
patterns while deviating substantially from its
temporal references.
We quantify this local disaffinity by measuring
its Euclidean distance to the mean fingerprint
of the non-overlapping temporal reference patches:
\begin{equation}
    S_i^{\mathrm{local}}
    =
    \left\|
        \tilde R_i
        -
        \frac{1}{n_i}
        \sum_{j\in\mathcal{N}_i}
        \tilde R_j
    \right\|_2,
    \label{eq:local_refi_score}
\end{equation}
where $\mathcal{N}_i$ denotes the non-overlapping
temporal reference set defined in
Eq.~\eqref{eq:refi_reference_set},
and $n_i=|\mathcal{N}_i|$.
The contextual dimensions $r_{i,1}$ and $r_{i,2}$
describe deviations relative to temporal references
during fingerprint construction, whereas this score
measures the discrepancy between complete ReFi
descriptors and their local reference mean.
A larger score indicates greater local statistical
inconsistency, even when the query remains globally
familiar.

\textbf{\textit{Two-Stage Score Fusion.}}
The three anomaly scores are integrated
through a two-stage fusion strategy.
The first stage combines deep-space discrepancy
and global statistical rarity, while the second
incorporates local contextual disaffinity.
Let $S^{\mathrm{lof}}$,
$S^{\mathrm{global}}$,
and $S^{\mathrm{local}}\in\mathbb{R}^{N}$
denote the patch-level score vectors for a test
time series $X^{te}$ containing $N$ patches.
The fusion procedure is defined as:
\begin{equation}
\begin{aligned}
   S^{\mathrm{ref}}
    &=
    \mathcal{Z}(S^{\mathrm{lof}})
    +
    \gamma_g
    \mathcal{Z}(S^{\mathrm{global}}), \\
   S^{\mathrm{patch}}
    &=
    \mathcal{Z}(S^{\mathrm{ref}})
    +
    \gamma_l
    \mathcal{Z}(S^{\mathrm{local}}),
\end{aligned}
\label{eq:sequential_score_fusion}
\end{equation}
where $S^{\mathrm{patch}}$ is the patch-level anomaly scores of $X^{te}$, 
and $\gamma_g, \gamma_l\geq0$
control the contributions of global and local statistical evidence, respectively.
The standardization operator
$\mathcal{Z}(\cdot)$ is defined as:
\begin{equation}
    \mathcal{Z}(\mathbf{a})
    =
    \begin{cases}
        \dfrac{
            \mathbf{a}-\bar a\mathbf{1}
        }{
            \operatorname{std}_0(\mathbf{a})
        },
        & \operatorname{std}_0(\mathbf{a})
        \geq\varepsilon_Z,\\
        \mathbf{0},
        & \text{otherwise},
    \end{cases}
    \label{eq:score_standardization}
\end{equation}
where $\bar a$ and $\operatorname{std}_0(\mathbf{a})$ denote the mean and population standard deviation of a score vector
$\mathbf{a}\in\mathbb{R}^{N}$,
respectively, and $\varepsilon_Z>0$
is a numerical threshold.
The second stage re-standardizes the combined
reference-based evidence before incorporating
local contextual deviations.
Finally, overlapping patch scores are aggregated
into point-wise anomaly scores.
For each test time step $t$, let
$\mathcal{P}_t^{\mathrm{te}}
=
\{P_i^{\mathrm{te}}\in\mathcal{P}^{\mathrm{te}}
\mid i\leq t\leq i+W-1\}$
denote the set of test patches covering $t$.
The corresponding point-wise anomaly score is
computed by averaging their patch-level scores
\cite{park2026paano}:
\begin{equation}
    s_t
    =
    \frac{1}{|\mathcal{P}_t^{\mathrm{te}}|}
    \sum_{P_i^{\mathrm{te}}\in\mathcal{P}_t^{\mathrm{te}}}
    S_i^{\mathrm{patch}},
    \qquad
    t=0,\ldots,T-1.
    \label{eq:pointwise_score_aggregation}
\end{equation}
The resulting point-wise anomaly score sequence is $\mathcal{S}^{\mathrm{te}}=(s_0,\ldots,s_{T-1})^\top \in\mathbb{R}^{T}$.

\begin{table}[]
\centering
\caption{Statistics of the TSB-AD benchmark.}
\label{tab:dataset_statistics}
\renewcommand{\arraystretch}{0.8}
\vspace{-0.3cm}
\resizebox{\columnwidth}{!}{%
\begin{tabular}{ccccc}
\toprule
\textbf{Category} & \textbf{Split} & \textbf{\# Time-Series}
& \textbf{Avg. Length} & \textbf{Anomaly Ratio} \\
\midrule
\multirow{2}{*}{TSB-AD-U}
& Tuning & 48  & 47143.3  & 3.5\% \\
& Eval   & 350 & 51866.7  & 4.5\% \\
\midrule
\multirow{2}{*}{TSB-AD-M}
& Tuning & 20  & 98164.1  & 5.7\% \\
& Eval   & 180 & 108826.7 & 5.0\% \\
\bottomrule
\end{tabular}}
\end{table}

\begin{table*}[]
\centering
\caption{Experimental results on \textbf{univariate time-series anomaly detection}
in TSB-AD-U. The best and second-best values are highlighted in
\textbf{bold} and \underline{underlined}, respectively.
Detection scores are reported as Score/Rank, with ties assigned
the same rank at the displayed precision.
Gain denotes the percentage-point ($\%$) improvement over the best baseline.}
\label{tab:comparison_u}
\vspace{-0.3cm}
\renewcommand{\arraystretch}{0.9}
\resizebox{0.85\textwidth}{!}{%
\begin{tabular}{c|l*{6}{l}rr}
\toprule
& \textbf{Method} & VUS-PR & VUS-ROC & Range-F1
& AUC-PR & AUC-ROC & Point-F1 & \# Params & Run Time \\

\midrule
\multirow{5}{*}{\rotatebox[origin=c]{90}{\textbf{Stat. \& ML}}}
& (Sub)-PCA~\cite{aggarwal2016introduction}
& 0.42\textsubscript{/6} & 0.76\textsubscript{/11}
& 0.41\textsubscript{/6} & 0.37\textsubscript{/7}
& 0.71\textsubscript{/9} & 0.42\textsubscript{/7}
& -- & \underline{1.5s} \\

& KShapeAD~\cite{paparrizos2015k}
& 0.40\textsubscript{/8} & 0.76\textsubscript{/11}
& 0.40\textsubscript{/8} & 0.35\textsubscript{/9}
& 0.74\textsubscript{/8} & 0.39\textsubscript{/9}
& -- & 8.0s \\

& DLinear~\cite{zamanzadeh2024deep}
& 0.27\textsubscript{/18} & 0.76\textsubscript{/11}
& 0.24\textsubscript{/18} & 0.23\textsubscript{/17}
& 0.67\textsubscript{/14} & 0.28\textsubscript{/17}
& \textbf{$<0.1$M} & 2.9s \\

& NLinear~\cite{zamanzadeh2024deep}
& 0.26\textsubscript{/20} & 0.74\textsubscript{/17}
& 0.21\textsubscript{/20} & 0.20\textsubscript{/20}
& 0.63\textsubscript{/20} & 0.24\textsubscript{/20}
& \textbf{$<0.1$M} & 5.8s \\

& MMPAD~\cite{yeh2026matrix}
& 0.41\textsubscript{/7} & 0.78\textsubscript{/8}
& 0.40\textsubscript{/8} & 0.37\textsubscript{/7}
& 0.76\textsubscript{/7} & 0.42\textsubscript{/7}
& -- & 61.8s \\

\midrule
\multirow{9}{*}{\rotatebox[origin=c]{90}{\textbf{NN}}}
& DeepAnT~\cite{munir2018deepant}
& 0.34\textsubscript{/13} & 0.79\textsubscript{/7}
& 0.35\textsubscript{/12} & 0.33\textsubscript{/10}
& 0.71\textsubscript{/9} & 0.38\textsubscript{/10}
& \textbf{$<0.1$M} & 2.0s \\

& USAD~\cite{audibert2020usad}
& 0.36\textsubscript{/12} & 0.71\textsubscript{/22}
& 0.40\textsubscript{/8} & 0.32\textsubscript{/11}
& 0.66\textsubscript{/17} & 0.37\textsubscript{/11}
& \textbf{$<0.1$M} & 1.7s \\

& TimesNet~\cite{wu2022timesnet}
& 0.26\textsubscript{/20} & 0.72\textsubscript{/20}
& 0.21\textsubscript{/20} & 0.18\textsubscript{/21}
& 0.61\textsubscript{/21} & 0.24\textsubscript{/20}
& \textbf{$<0.1$M} & 11.2s \\

& FITS~\cite{xu2024fits}
& 0.26\textsubscript{/20} & 0.73\textsubscript{/19}
& 0.20\textsubscript{/22} & 0.17\textsubscript{/22}
& 0.61\textsubscript{/21} & 0.23\textsubscript{/22}
& \textbf{$<0.1$M} & 3.1s \\

& DADA~\cite{shentu2025towards}
& 0.31\textsubscript{/15} & 0.77\textsubscript{/9}
& 0.31\textsubscript{/15} & 0.29\textsubscript{/14}
& 0.71\textsubscript{/9} & 0.33\textsubscript{/15}
& 1.84M & \textbf{0.8s} \\

& KAN-AD~\cite{pmlr-v267-zhou25u}
& 0.40\textsubscript{/8} & 0.82\textsubscript{/6}
& 0.43\textsubscript{/5} & 0.41\textsubscript{/6}
& 0.80\textsubscript{/6} & 0.44\textsubscript{/6}
& \textbf{$<0.1$M} & 12.1s \\

& CoAD~\cite{tang2026bridging}
& 0.53\textsubscript{/3} & 0.88\textsubscript{/4}
& 0.49\textsubscript{/3} & 0.48\textsubscript{/3}
& 0.85\textsubscript{/4} & 0.52\textsubscript{/3}
& 0.78M & 154.6s \\

& PaAno~\cite{park2026paano}
& 0.53\textsubscript{/3} & 0.89\textsubscript{/3}
& 0.49\textsubscript{/3} & 0.47\textsubscript{/4}
& \underline{0.87}\textsubscript{/2} & 0.52\textsubscript{/3}
& \underline{0.3M} & 5.4s \\

& PaAno+~\cite{zhu2026paano+}
& \underline{0.55}\textsubscript{/2}
& \underline{0.90}\textsubscript{/2}
& \underline{0.52}\textsubscript{/2}
& \underline{0.49}\textsubscript{/2}
& \underline{0.87}\textsubscript{/2}
& \underline{0.54}\textsubscript{/2}
& 1.1M & 12.9s \\

\midrule
\multirow{10}{*}{\rotatebox[origin=c]{90}{\textbf{Transformer}}}
& AnomalyTransformer~\cite{xu2022anomalytransformer}
& 0.12\textsubscript{/24} & 0.56\textsubscript{/25}
& 0.14\textsubscript{/24} & 0.08\textsubscript{/24}
& 0.50\textsubscript{/24} & 0.12\textsubscript{/24}
& 4.8M & 48.9s \\

& DCdetector~\cite{yang2023dcdetector}
& 0.09\textsubscript{/25} & 0.57\textsubscript{/24}
& 0.07\textsubscript{/25} & 0.05\textsubscript{/25}
& 0.50\textsubscript{/24} & 0.10\textsubscript{/25}
& 0.8M & 5.8s \\

& Lag-Llama~\cite{rasul2023lag}
& 0.27\textsubscript{/18} & 0.72\textsubscript{/20}
& 0.31\textsubscript{/15} & 0.25\textsubscript{/16}
& 0.65\textsubscript{/18} & 0.30\textsubscript{/16}
& 2.5M & 1220.8s \\

& OFA~\cite{zhou2023one}
& 0.24\textsubscript{/23} & 0.71\textsubscript{/22}
& 0.20\textsubscript{/22} & 0.16\textsubscript{/23}
& 0.59\textsubscript{/23} & 0.22\textsubscript{/23}
& 81.9M & 171.1s \\

& PatchTST~\cite{nie2023a}
& 0.30\textsubscript{/16} & 0.76\textsubscript{/11}
& 0.25\textsubscript{/17} & 0.23\textsubscript{/17}
& 0.65\textsubscript{/18} & 0.27\textsubscript{/18}
& 6.9M & 26.3s \\

& iTransformer~\cite{liu2024itransformer}
& 0.32\textsubscript{/14} & 0.77\textsubscript{/9}
& 0.22\textsubscript{/19} & 0.21\textsubscript{/19}
& 0.67\textsubscript{/14} & 0.26\textsubscript{/19}
& 6.3M & 9.8s \\

& MOMENT (FT)~\cite{goswami2024moment}
& 0.39\textsubscript{/10} & 0.76\textsubscript{/11}
& 0.35\textsubscript{/12} & 0.30\textsubscript{/12}
& 0.69\textsubscript{/12} & 0.35\textsubscript{/12}
& 109.6M & 43.6s \\

& MOMENT (ZS)~\cite{goswami2024moment}
& 0.38\textsubscript{/11} & 0.75\textsubscript{/16}
& 0.36\textsubscript{/11} & 0.30\textsubscript{/12}
& 0.68\textsubscript{/13} & 0.35\textsubscript{/12}
& 109.6M & 42.9s \\

& TimesFM~\cite{das2024decoder}
& 0.30\textsubscript{/16} & 0.74\textsubscript{/17}
& 0.34\textsubscript{/14} & 0.28\textsubscript{/15}
& 0.67\textsubscript{/14} & 0.34\textsubscript{/14}
& 203.5M & 83.8s \\

& TimeRCD~\cite{lan2026timercd}
& 0.46\textsubscript{/5} & 0.85\textsubscript{/5}
& 0.41\textsubscript{/6} & 0.44\textsubscript{/5}
& 0.82\textsubscript{/5} & 0.49\textsubscript{/5}
& 37.1M & -- \\

\midrule
\rowcolor{cmpOursBg}
& ReFi-TS (Ours)
& \textbf{0.62\textsubscript{/1}}
& \textbf{0.93\textsubscript{/1}}
& \textbf{0.57\textsubscript{/1}}
& \textbf{0.56\textsubscript{/1}}
& \textbf{0.91\textsubscript{/1}}
& \textbf{0.60\textsubscript{/1}}
& 0.39M & 5.76s \\

\rowcolor{cmpGainBg}
& \textit{Gain ($\%$)}
& +7.00 & +3.00 & +5.00
& +7.00 & +4.00 & +6.00
& -- & -- \\
\bottomrule
\end{tabular}}
\end{table*}

\section{Experiments}
\subsection{Experimental Setup}
\textbf{\textit{a) Datasets.}}
We evaluate ReFi-TS on TSB-AD~\cite{liu2024elephant}, 
a large-scale time-series anomaly detection benchmark comprising 
870 univariate (TSB-AD-U) and 200 multivariate (TSB-AD-M) time series 
curated from 40 datasets.
 The summary
statistics are presented in Table \ref{tab:dataset_statistics}.
TSB-AD covers diverse application domains, including 
industrial monitoring, web services, transportation, finance, 
healthcare, and environmental monitoring, 
with varying temporal characteristics and anomaly patterns.
Following PaAno \cite{park2026paano} and CoAD \cite{tang2026bridging}, 
we adopt the predefined training and evaluation splits, 
which contain 48 and 350 time series for TSB-AD-U 
and 20 and 180 time series for TSB-AD-M, respectively.
For each time series, ReFi-TS is fitted exclusively on 
the predefined training segment, without access to anomaly labels.
All model parameters, fingerprint normalization statistics, 
normal-mode prototypes, and reference banks are derived 
solely from this segment.

\textbf{\textit{b) Baselines.}}
We compare ReFi-TS with a comprehensive set of representative 
baselines and recent state-of-the-art methods for univariate 
and multivariate time series anomaly detection, which are 
categorized into three groups:
(i) \textit{Statistical and machine learning methods}, including 
KMeansAD~\cite{yairi2001fault}, 
(Sub)-PCA~\cite{aggarwal2016introduction}, 
KShapeAD~\cite{paparrizos2015k}, 
DLinear and NLinear~\cite{zamanzadeh2024deep}, 
and MMPAD~\cite{yeh2026matrix};
(ii) \textit{Neural network-based methods}, including 
DeepAnT~\cite{munir2018deepant}, 
OmniAnomaly~\cite{su2019robust}, 
USAD~\cite{audibert2020usad}, 
TimesNet~\cite{wu2022timesnet}, 
FITS~\cite{xu2024fits}, 
DADA~\cite{shentu2025towards}, 
KAN-AD~\cite{pmlr-v267-zhou25u}, 
CoAD~\cite{tang2026bridging}, 
STREAM-VAE~\cite{ozer2026stream}, 
TSPulse~\cite{ekambaram2026tspulse}, 
PaAno~\cite{park2026paano}, 
and PaAno+~\cite{zhu2026paano+};
(iii) \textit{Transformer-based methods}, including 
AnomalyTransformer~\cite{xu2022anomalytransformer}, 
DCdetector~\cite{yang2023dcdetector}, 
Lag-Llama~\cite{rasul2023lag}, 
OFA~\cite{zhou2023one}, 
PatchTST~\cite{nie2023a}, 
iTransformer~\cite{liu2024itransformer}, 
MOMENT~\cite{goswami2024moment}, 
TimesFM~\cite{das2024decoder}, 
CATCH~\cite{wu2025catch}, 
and TimeRCD~\cite{lan2026timercd}.
Both MOMENT and TSPulse are evaluated under zero-shot (ZS) 
and fine-tuning (FT) settings.

\textbf{\textit{c) Implementation Details.}}
The patch encoder consists of four 1D convolutional layers
followed by global average pooling, producing 64-dimensional embeddings.
The projector is a two-layer MLP with an output
dimension of 256.
The ReFi encoder comprises two parallel two-layer MLPs,
each producing 64-dimensional embeddings, followed by
a learnable gating mechanism.
The Comparator is implemented as a two-layer MLP
with a hidden dimension of 64.

ReFi-TS is trained independently for each time series
using AdamW for 100 iterations with a mini-batch size
of 512 and a weight decay of $10^{-4}$.
The learning rate follows a cosine decay schedule
from $10^{-4}$ to $10^{-5}$.
The patch length $W$ is set to 96 with a stride of 1,
and increased to 192 for training sequences exceeding
30,000 time points.
The number of normal modes is set to $K_m=10$.
The ReFi neighborhood size $h$ is set to 60 for
TSB-AD-U and 50 for TSB-AD-M.
For anomaly scoring, the deep representation and ReFi
reference banks are constructed using sampling ratios
of 10\% and 20\%, respectively.
The numbers of nearest neighbors in
$\mathcal{K}_H$ and $\mathcal{K}_R$ are both set to 3.
The score fusion coefficients are set to
$\gamma_g=0.8$ for both benchmarks, while
$\gamma_l=0.3$ for TSB-AD-U and
$\gamma_l=0.2$ for TSB-AD-M.
The loss weights
$\lambda_1$, $\lambda_2$, $\lambda_3$, and $\lambda_4$
are set to 1, 0.5, 0.5, and 0.01, respectively.
They are selected via grid search over $[0.001,1]$,
following established practices~\cite{lu2024caudits,ozyurt2023contrastive}.
 
\textbf{\textit{d) Evaluation Metrics.}}
Following the TSB-AD evaluation protocol~\cite{liu2024elephant},
we employ six complementary metrics, including three
range-wise metrics (VUS-PR and VUS-ROC~\cite{paparrizos2022volume},
and Range-wise F1 score (Range-F1)) and three point-wise
metrics (AUC-PR~\cite{davis2006relationship},
AUC-ROC~\cite{park2026paano}, and
Point-F1~\cite{sarfraz2024icml}).
VUS-PR serves as the primary metric owing to its robustness
to temporal misalignment and its reliability under class
imbalance~\cite{paparrizos2022volume,liu2024elephant}.
We additionally report trainable parameters and runtime 
to assess computational efficiency.
Each experiment is repeated 10 times with different
random seeds, and the average results are reported.

\begin{table*}[!t]
\centering
\caption{Experimental results on \textbf{multivariate time-series anomaly detection}
in TSB-AD-M.
The best and second-best comparable values are highlighted in
\textbf{bold} and \underline{underlined}, respectively.
Detection scores are reported as Score/Rank.
The final row reports the gain in percentage points ($\%$)
over the best comparable baseline.}
\label{tab:comparison_m}
\renewcommand{\arraystretch}{0.9}
\providecommand{\tsbadscore}[2]{#1\textsubscript{/#2}}
\vspace{-0.3cm}
\resizebox{0.85\textwidth}{!}{%
\begin{tabular}{c|@{\hspace{3pt}}l*{8}{c}}
\toprule
& \textbf{Method} & VUS-PR & VUS-ROC
& Range-F1 & AUC-PR & AUC-ROC & Point-F1
& \# Params & Run Time \\
\midrule
\multirow{5}{*}{\rotatebox[origin=c]{90}{\bfseries Stat.\&ML}}
& KMeansAD~\cite{yairi2001fault}
& \tsbadscore{0.29}{18} & \tsbadscore{0.73}{11}
& \tsbadscore{0.33}{10} & \tsbadscore{0.25}{17}
& \tsbadscore{0.69}{11} & \tsbadscore{0.31}{16}
& -- & 62.0s \\

& PCA~\cite{aggarwal2016introduction}
& \tsbadscore{0.31}{10} & \tsbadscore{0.74}{9}
& \tsbadscore{0.29}{11} & \tsbadscore{0.31}{9}
& \tsbadscore{0.70}{9} & \tsbadscore{0.37}{8}
& -- & \textbf{0.1s} \\

& DLinear~\cite{zamanzadeh2024deep}
& \tsbadscore{0.29}{18} & \tsbadscore{0.70}{17}
& \tsbadscore{0.21}{16} & \tsbadscore{0.27}{14}
& \tsbadscore{0.66}{15} & \tsbadscore{0.32}{14}
& \textbf{$<0.1$M} & 14.8s \\

& NLinear~\cite{zamanzadeh2024deep}
& \tsbadscore{0.30}{14} & \tsbadscore{0.70}{17}
& \tsbadscore{0.21}{16} & \tsbadscore{0.26}{16}
& \tsbadscore{0.65}{16} & \tsbadscore{0.31}{16}
& \textbf{$<0.1$M} & 15.0s \\

& MMPAD~\cite{yeh2026matrix}
& \tsbadscore{0.35}{9} & \tsbadscore{0.75}{8}
& \tsbadscore{0.29}{11} & \tsbadscore{0.30}{12}
& \tsbadscore{0.74}{5} & \tsbadscore{0.36}{11}
& -- & 690.5s \\

\midrule
\multirow{11}{*}{\rotatebox[origin=c]{90}{\bfseries NN}}
& DeepAnT~\cite{munir2018deepant}
& \tsbadscore{0.31}{10} & \tsbadscore{0.76}{5}
& \tsbadscore{0.37}{8} & \tsbadscore{0.32}{8}
& \tsbadscore{0.73}{6} & \tsbadscore{0.37}{8}
& \textbf{$<0.1$M} & 9.5s \\

& OmniAnomaly~\cite{su2019robust}
& \tsbadscore{0.31}{10} & \tsbadscore{0.69}{19}
& \tsbadscore{0.37}{8} & \tsbadscore{0.27}{14}
& \tsbadscore{0.65}{16} & \tsbadscore{0.32}{14}
& \textbf{$<0.1$M} & 9.1s \\

& TimesNet~\cite{wu2022timesnet}
& \tsbadscore{0.19}{22} & \tsbadscore{0.64}{21}
& \tsbadscore{0.17}{20} & \tsbadscore{0.13}{22}
& \tsbadscore{0.56}{21} & \tsbadscore{0.20}{22}
& \textbf{$<0.1$M} & 52.1s \\

& FITS~\cite{xu2024fits}
& \tsbadscore{0.21}{20} & \tsbadscore{0.66}{20}
& \tsbadscore{0.16}{22} & \tsbadscore{0.15}{20}
& \tsbadscore{0.58}{20} & \tsbadscore{0.22}{20}
& \textbf{$<0.1$M} & 16.7s \\

& DADA~\cite{shentu2025towards}
& \tsbadscore{0.31}{10} & \tsbadscore{0.73}{11}
& \tsbadscore{0.25}{13} & \tsbadscore{0.31}{9}
& \tsbadscore{0.69}{11} & \tsbadscore{0.35}{12}
& 1.84M & \underline{2.1s} \\

& KAN-AD~\cite{pmlr-v267-zhou25u}
& \tsbadscore{0.40}{6} & \tsbadscore{0.76}{5}
& \tsbadscore{0.41}{5} & \tsbadscore{0.35}{6}
& \tsbadscore{0.73}{6} & \tsbadscore{0.40}{6}
& \textbf{$<0.1$M} & 31.9s \\

& STREAM-VAE~\cite{ozer2026stream}
& \tsbadscore{\underline{0.44}}{2}
& \tsbadscore{0.78}{4}
& \tsbadscore{\underline{0.42}}{2}
& \tsbadscore{\underline{0.38}}{2}
& \tsbadscore{0.76}{3}
& \tsbadscore{\underline{0.45}}{2}
& 2.69M & -- \\

& TSPulse (ZS)~\cite{ekambaram2026tspulse}
& \tsbadscore{0.40}{6}
& \tsbadscore{0.72}{13}
& \tsbadscore{\underline{0.42}}{2}
& \tsbadscore{0.35}{6}
& \tsbadscore{0.67}{13}
& \tsbadscore{0.40}{6}
& 1.0M & -- \\

& TSPulse (FT)~\cite{ekambaram2026tspulse}
& \tsbadscore{\underline{0.44}}{2}
& \tsbadscore{0.76}{5}
& \tsbadscore{\underline{0.42}}{2}
& \tsbadscore{\underline{0.38}}{2}
& \tsbadscore{0.72}{8}
& \tsbadscore{0.43}{3}
& 1.0M & -- \\

& PaAno~\cite{park2026paano}
& \tsbadscore{0.43}{4}
& \tsbadscore{0.79}{3}
& \tsbadscore{0.41}{5}
& \tsbadscore{\underline{0.38}}{2}
& \tsbadscore{0.76}{3}
& \tsbadscore{0.43}{3}
& \underline{0.3M} & 9.7s \\

& PaAno+~\cite{zhu2026paano+}
& \tsbadscore{0.42}{5}
& \tsbadscore{\underline{0.80}}{2}
& \tsbadscore{0.41}{5}
& \tsbadscore{\underline{0.38}}{2}
& \tsbadscore{\underline{0.77}}{2}
& \tsbadscore{0.43}{3}
& 1.5M & 23.5s \\

\midrule
\multirow{7}{*}{\rotatebox[origin=c]{90}{\bfseries Transformer}}
& AnomalyTransformer~\cite{xu2022anomalytransformer}
& \tsbadscore{0.12}{23} & \tsbadscore{0.57}{23}
& \tsbadscore{0.14}{23} & \tsbadscore{0.07}{23}
& \tsbadscore{0.52}{23} & \tsbadscore{0.12}{23}
& 4.8M & 55.8s \\

& DCdetector~\cite{yang2023dcdetector}
& \tsbadscore{0.10}{24} & \tsbadscore{0.54}{24}
& \tsbadscore{0.09}{24} & \tsbadscore{0.05}{24}
& \tsbadscore{0.48}{24} & \tsbadscore{0.10}{24}
& 0.8M & 15.0s \\

& OFA~\cite{zhou2023one}
& \tsbadscore{0.21}{20} & \tsbadscore{0.63}{22}
& \tsbadscore{0.17}{20} & \tsbadscore{0.15}{20}
& \tsbadscore{0.55}{22} & \tsbadscore{0.21}{21}
& 81.9M & 532.9s \\

& PatchTST~\cite{nie2023a}
& \tsbadscore{0.30}{14} & \tsbadscore{0.71}{15}
& \tsbadscore{0.22}{15} & \tsbadscore{0.28}{13}
& \tsbadscore{0.67}{13} & \tsbadscore{0.33}{13}
& 6.9M & 66.9s \\

& iTransformer~\cite{liu2024itransformer}
& \tsbadscore{0.30}{14} & \tsbadscore{0.71}{15}
& \tsbadscore{0.18}{18} & \tsbadscore{0.24}{18}
& \tsbadscore{0.64}{19} & \tsbadscore{0.29}{18}
& 6.4M & 24.4s \\

& CATCH~\cite{wu2025catch}
& \tsbadscore{0.30}{14} & \tsbadscore{0.72}{13}
& \tsbadscore{0.18}{18} & \tsbadscore{0.23}{19}
& \tsbadscore{0.65}{16} & \tsbadscore{0.29}{18}
& 210.8M & 40.1s \\

& TimeRCD~\cite{lan2026timercd}
& \tsbadscore{0.36}{8}
& \tsbadscore{0.74}{9}
& \tsbadscore{0.25}{13}
& \tsbadscore{0.31}{9}
& \tsbadscore{0.70}{9}
& \tsbadscore{0.37}{8}
& 37.1M & -- \\

\midrule
\rowcolor{cmpOursBg}
&
\textbf{ReFi-TS (Ours)}
& \textbf{\tsbadscore{0.47}{1}}
& \textbf{\tsbadscore{0.82}{1}}
& \textbf{\tsbadscore{0.45}{1}}
& \textbf{\tsbadscore{0.43}{1}}
& \textbf{\tsbadscore{0.80}{1}}
& \textbf{\tsbadscore{0.46}{1}}
& 0.42M & 11.42s \\

\rowcolor{cmpGainBg}
&
\textit{Gain ($\%$)}
& 3.00 & 2.00 & 3.00
& 5.00 & 3.00 & 1.00
& -- & -- \\

\bottomrule
\end{tabular}%
}
\end{table*}

\begin{figure*}[t]
    \centering
    \includegraphics[width=\textwidth]{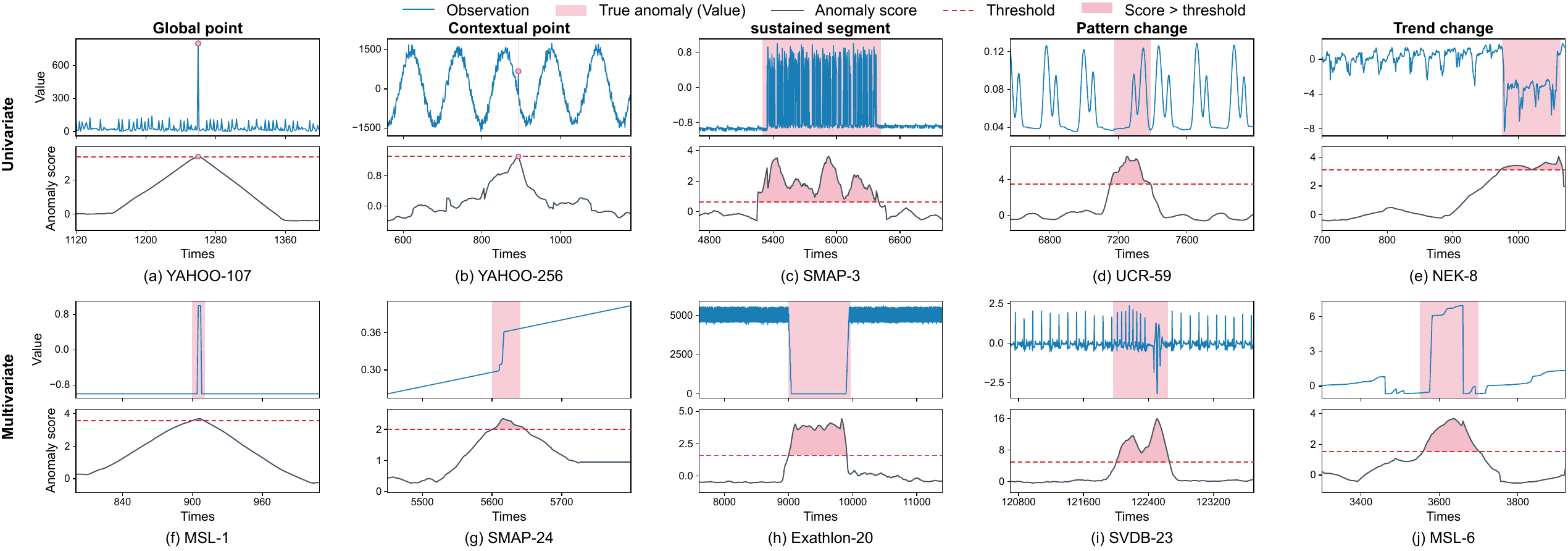}
    \vspace{-0.7cm}
   \caption{Qualitative anomaly detection results on ten 
        representative cases from univariate and multivariate 
        time series. Pink regions highlight ground-truth and 
        detected anomalies, respectively.}
    \label{fig:qualitative}
\end{figure*}

\begin{figure}[!t]
    \centering
    \includegraphics[width=0.9\columnwidth]
    {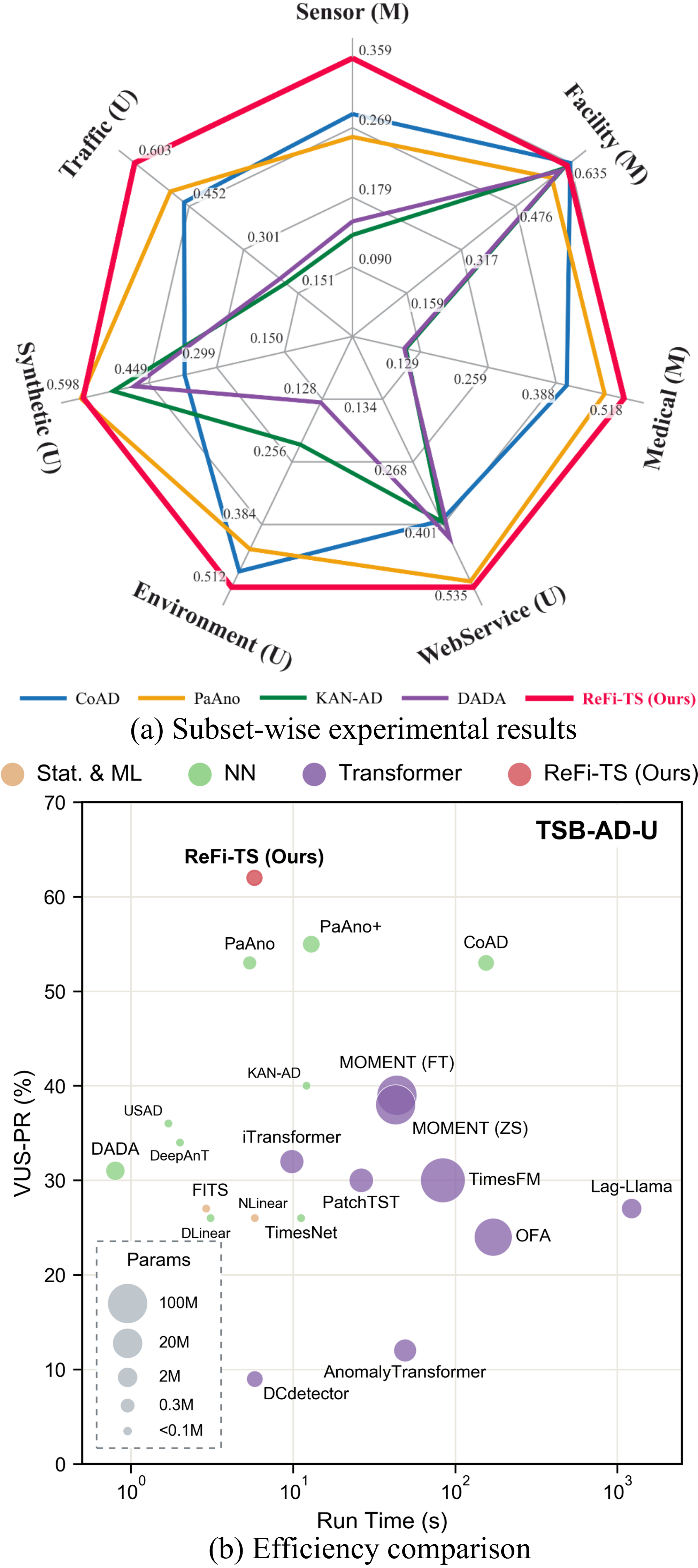}
    \vspace{-0.3cm}
   \caption{Subset-wise performance comparison on TSB-AD dataset, and efficiency comparison on TSB-AD-U.}
    \label{fig:efficiency}
\end{figure}

\begin{table}[]
\centering
\caption{Effectiveness of individual loss components on
TSB-AD-U and TSB-AD-M.}
\label{tab:loss_ablation}
\renewcommand{\arraystretch}{0.85}
\vspace{-0.3cm}
\resizebox{\columnwidth}{!}{%
\begin{tabular}
{@{\extracolsep{\fill}}l*{6}{c}@{}}
\toprule
Variant
& \shortstack{VUS-PR}
& \shortstack{VUS-ROC}
& \shortstack{Range-F1}
& \shortstack{AUC-PR}
& \shortstack{AUC-ROC}
& \shortstack{Point-F1} \\
\midrule
\multicolumn{7}{l}{\textbf{TSB-AD-M}} \\
\addlinespace[2pt]

w/o BA
& 46.10 & 81.90 & 44.83 & 42.33 & 79.81 & 45.05 \\

w/o RM
& 45.91 & 81.73 & 43.41 & 42.12 & 79.60 & 44.55 \\

w/o SMCH
& 46.62 & 82.07 & 44.90 & 42.48 & 79.90 & 45.52 \\

w/o CON
& 47.00 & 82.00 & \textbf{45.00}
& 43.00 & 80.00 & 46.00 \\

Full
& \textbf{47.14} & \textbf{82.12} & 44.73
& \textbf{43.13} & \textbf{80.02} & \textbf{46.27} \\

\midrule

\multicolumn{7}{l}{\textbf{TSB-AD-U}} \\
\addlinespace[2pt]

w/o BA
& 62.00 & \textbf{93.00} & 57.00
& 56.00 & 91.00 & 60.00 \\

w/o RM
& 61.92 & 92.97 & 56.50
& 55.70 & 90.91 & 59.64 \\

w/o SMCH
& 61.81 & 92.85 & 56.77
& 56.19 & 90.85 & 60.02 \\

w/o CON
& 61.59 & 92.63 & 56.40
& 55.89 & 90.64 & 59.61 \\

Full
& \textbf{62.54} & 92.99 & \textbf{57.08}
& \textbf{56.49} & \textbf{91.12} & \textbf{60.30} \\

\bottomrule
\end{tabular}}
\vspace{2pt}
\parbox{\columnwidth}{\footnotesize
\textit{Note:} BA, RM, SMCH, and CON denote
$\mathcal{L}_{\mathrm{ba}}$,
$\mathcal{L}_{\mathrm{rm}}$,
$\mathcal{L}_{\mathrm{smch}}$, and
$\mathcal{L}_{\mathrm{con}}$, respectively.
}
\end{table}

\subsection{Results}
\textbf{\textit{a) Comparison with Baselines.}}
Tables~\ref{tab:comparison_u} and~\ref{tab:comparison_m}
report the comparative results on TSB-AD-U and TSB-AD-M,
respectively.
ReFi-TS achieves the highest scores across all six
evaluation metrics on both benchmarks, demonstrating
consistent performance under univariate and multivariate
settings.
On TSB-AD-U, it obtains VUS-PR and AUC-PR scores of
0.62 and 0.56, respectively, compared with 0.55 and
0.49 achieved by PaAno+~\cite{zhu2026paano+}.
On TSB-AD-M, ReFi-TS achieves corresponding scores
of 0.47 and 0.43, exceeding the best baseline results
of 0.44 and 0.38.
Similar advantages are observed in Range-F1, AUC-ROC,
and Point-F1, indicating its effectiveness in identifying
both point-wise anomalies and temporally extended
anomalous patterns.

Among conventional statistical methods,  MMPAD~\cite{yeh2026matrix} achieves
VUS-PR scores of 0.41 and 0.35 on the two benchmarks,
respectively.
Recent patch-based methods, particularly PaAno \cite{park2026paano} and
PaAno+ \cite{zhu2026paano+}, exhibit stronger competitive performance.
Nevertheless, ReFi-TS achieves VUS-PR scores of 0.62
and 0.47, substantially exceeding both methods on
TSB-AD-U and TSB-AD-M.
On the multivariate benchmark, STREAM-VAE \cite{ozer2026stream} and
TSPulse (FT)~\cite{ekambaram2026tspulse} achieve the second-best VUS-PR score
of 0.44, while ReFi-TS maintains a clear performance
advantage.
These results support the effectiveness of explicitly
incorporating contextual statistical deviations and
intrinsic temporal structures into deep representation
learning, complemented by mode-conditional normality
modeling.

\textbf{\textit{b) Qualitative Evaluation.}}
Figure~\ref{fig:qualitative} illustrates ten representative
cases spanning five anomaly categories, with univariate
and multivariate examples presented in the upper and lower
rows, respectively.
ReFi-TS exhibits distinct scoring patterns across anomaly
morphologies.
For global point anomalies (YAHOO-107 and MSL-1),
it produces localized peaks aligned with abrupt observations,
while contextual point anomalies (YAHOO-256 and SMAP-24)
trigger elevated responses despite their less distinctive
absolute magnitudes.
For sustained anomalies (SMAP-3 and Exathlon-20),
the scores remain elevated over the abnormal intervals
and recede as normal behavior resumes.
Pattern changes (UCR-59 and SVDB-23) induce multiple
pronounced responses associated with variations in waveform
and oscillatory structures.
For trend and level changes (NEK-8 and MSL-6),
the scores progressively increase or remain elevated
following persistent changes in temporal behavior.

These observations reveal that ReFi-TS adapts its scoring
responses to the underlying anomaly structure, rather than
relying solely on instantaneous magnitude deviations.
Its ability to capture localized deviations, temporal
persistence, and structural transitions is consistent
across both univariate and multivariate cases.
Furthermore, Figure \ref{fig:efficiency}(a) demonstrates its effectiveness
across six heterogeneous data subsets, where ReFi-TS
consistently achieves competitive or superior performance
compared with existing competitive baselines.
The complementary case-level and subset-wise results
support the robustness of ReFi-TS across diverse
temporal behaviors and data characteristics.

\textbf{\textit{c) Performance-Efficiency Trade-off.}}
Figure~\ref{fig:efficiency}(b) compares detection performance,
runtime, and model complexity on TSB-AD-U, with bubble
size indicating the number of parameters.
ReFi-TS achieves the highest VUS-PR of 62.0\% with
0.39M parameters and a runtime of 5.76s.
Compared with PaAno, it improves VUS-PR from 53.0\%
to 62.0\%, with a modest increase in model size and
runtime.
Relative to PaAno+, ReFi-TS achieves a 7-percentage-point
improvement in VUS-PR while reducing the parameter count
and runtime by 64.5\% and 55.3\%, respectively.
The advantage is particularly pronounced over CoAD,
which requires 154.6s to achieve a VUS-PR of 53.0\%,
whereas ReFi-TS attains substantially higher accuracy
with considerably lower runtime.

ReFi-TS also exhibits favorable efficiency compared
with Transformer-based methods.
MOMENT (FT) and MOMENT (ZS)~\cite{goswami2024moment} achieve VUS-PR scores
of 39.0\% and 38.0\%, respectively, with runtimes
exceeding 42s.
PatchTST~\cite{nie2023a} and TimesFM~\cite{das2024decoder} obtain VUS-PR scores of
approximately 30.0\%, requiring 26.3s and 83.8s,
respectively.
ReFi-TS achieves substantially higher detection
performance while reducing runtime by approximately
78\% to 93\% across these comparisons.
Overall, ReFi-TS provides a favorable trade-off between
detection accuracy, model size, and computational
efficiency, achieving the best detection performance
without relying on large-scale model capacity.

\begin{figure}[!t]
    \centering
    \includegraphics[width=0.85\columnwidth]{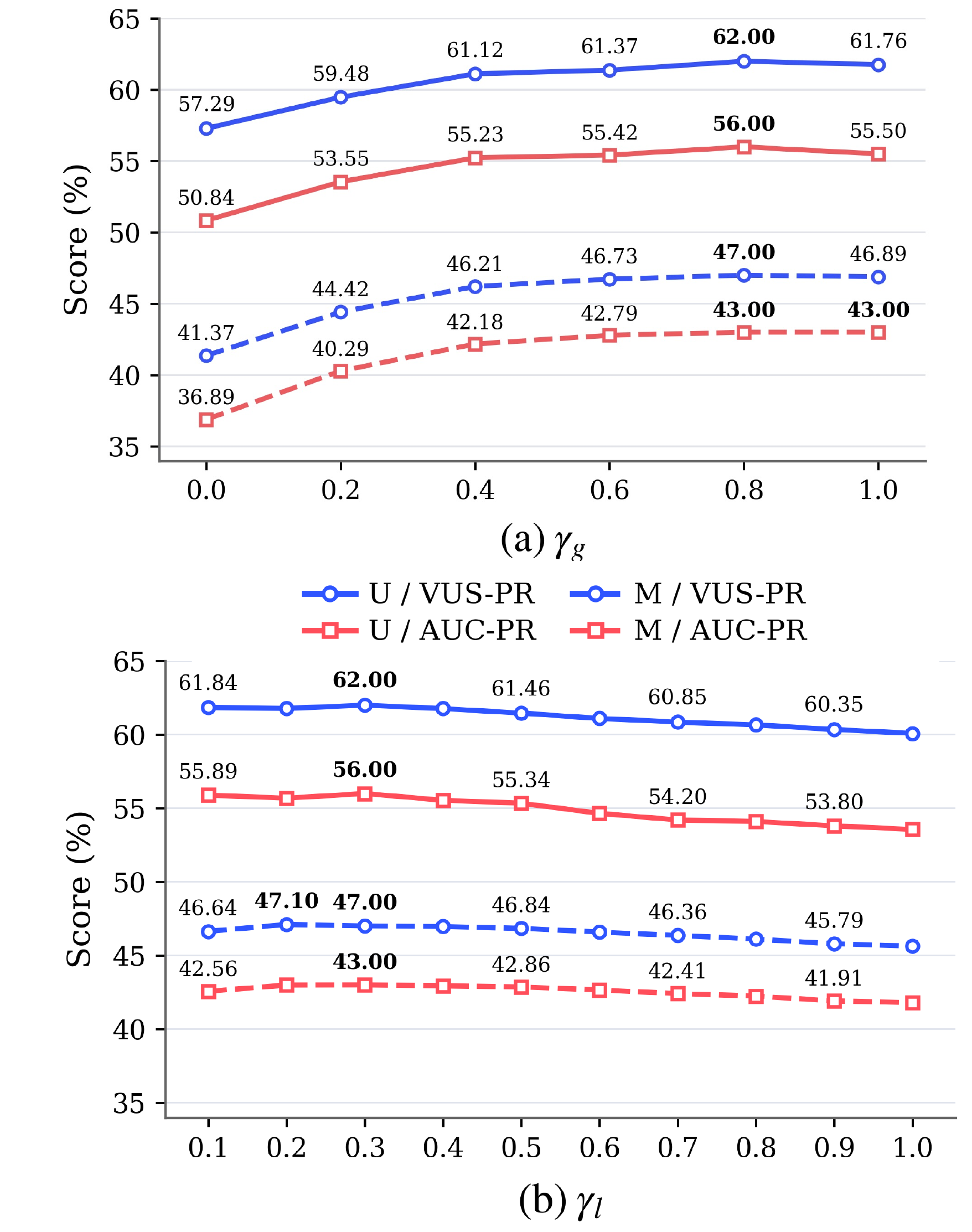}
    \vspace{-0.3cm}
    \caption{Sensitivity of  $\gamma_g$ and $\gamma_l$ on TSB-AD-U and
    TSB-AD-M. Performance is evaluated using VUS-PR
    and AUC-PR (\%).}
    \label{fig:gamma_sensitivity}
\end{figure}

\subsection{Ablation Study}
\textbf{\textit{a) Contribution of Each Component.}}
Table~\ref{tab:loss_ablation} evaluates the contribution
of the four loss components on TSB-AD-U and TSB-AD-M.
On TSB-AD-M, removing $\mathcal{L}_{\mathrm{rm}}$
causes the most substantial degradation across all six
metrics, with Range-F1 decreasing from 44.73\% to
43.41\% and Point-F1 from 46.27\% to 44.55\%.
Removing $\mathcal{L}_{\mathrm{ba}}$ also degrades
five metrics, particularly VUS-PR and Point-F1.
The removal of $\mathcal{L}_{\mathrm{smch}}$ primarily
affects Point-F1 and AUC-PR, which decrease from
46.27\% to 45.52\% and from 43.13\% to 42.48\%,
respectively.
In contrast, removing $\mathcal{L}_{\mathrm{con}}$
causes relatively minor changes.

On TSB-AD-U, $\mathcal{L}_{\mathrm{con}}$ plays
a more prominent role, as its removal leads to
the largest degradation across all six metrics,
with VUS-PR decreasing from 62.54\% to 61.59\%
and Point-F1 from 60.30\% to 59.61\%.
Removing $\mathcal{L}_{\mathrm{rm}}$ also 
degrades all six metrics, particularly AUC-PR,
which decreases from 56.49\% to 55.70\%.
The removal of $\mathcal{L}_{\mathrm{smch}}$ 
reduces all six metrics, whereas removing
$\mathcal{L}_{\mathrm{ba}}$ degrades five metrics,
with only a marginal improvement in VUS-ROC.

Overall, $\mathcal{L}_{\mathrm{rm}}$ provides consistent
benefits across both settings, while
$\mathcal{L}_{\mathrm{con}}$ contributes more
substantially to univariate detection.
The full objective achieves the best results on
five of the six metrics on each benchmark, with
minor trade-offs in Range-F1 on TSB-AD-M and
VUS-ROC on TSB-AD-U.
These results demonstrate the complementary
contributions of the four objectives to normality
representation learning.

\textbf{\textit{b) Sensitivity Analysis of  $\gamma_g$ and $\gamma_l$.}} 
Figure~\ref{fig:gamma_sensitivity} investigates the effects
of the fusion weights $\gamma_g$ and $\gamma_l$ on
TSB-AD-U and TSB-AD-M.
As shown in Fig.~\ref{fig:gamma_sensitivity}(a),
increasing $\gamma_g$ from 0 to 0.8 improves
VUS-PR by 4.71 and 5.63 percentage points on
TSB-AD-U and TSB-AD-M, respectively, while
AUC-PR increases by 5.16 and 6.11 points.
Performance  peaks at $\gamma_g=0.8$,
with no further improvement in VUS-PR at larger values.
For $\gamma_l$, Fig.~\ref{fig:gamma_sensitivity}(b)
shows that both datasets achieve favorable performance
around 0.2--0.4.
In particular, $\gamma_l=0.3$ yields the highest
VUS-PR and AUC-PR on TSB-AD-U, reaching
62.00\% and 56.00\%, respectively.
On TSB-AD-M, it achieves the highest AUC-PR of
43.00\% and a VUS-PR of 47.00\%, close to the
maximum of 47.10\% at $\gamma_l=0.2$.
Larger values of $\gamma_l$ gradually degrade
performance on both datasets.
Accordingly, we adopt $\gamma_g=0.8$ for both benchmarks, while
$\gamma_l=0.3$ for TSB-AD-U and
$\gamma_l=0.2$ for TSB-AD-M.

\begin{figure}[!t]
    \centering
    \includegraphics[width=\columnwidth]{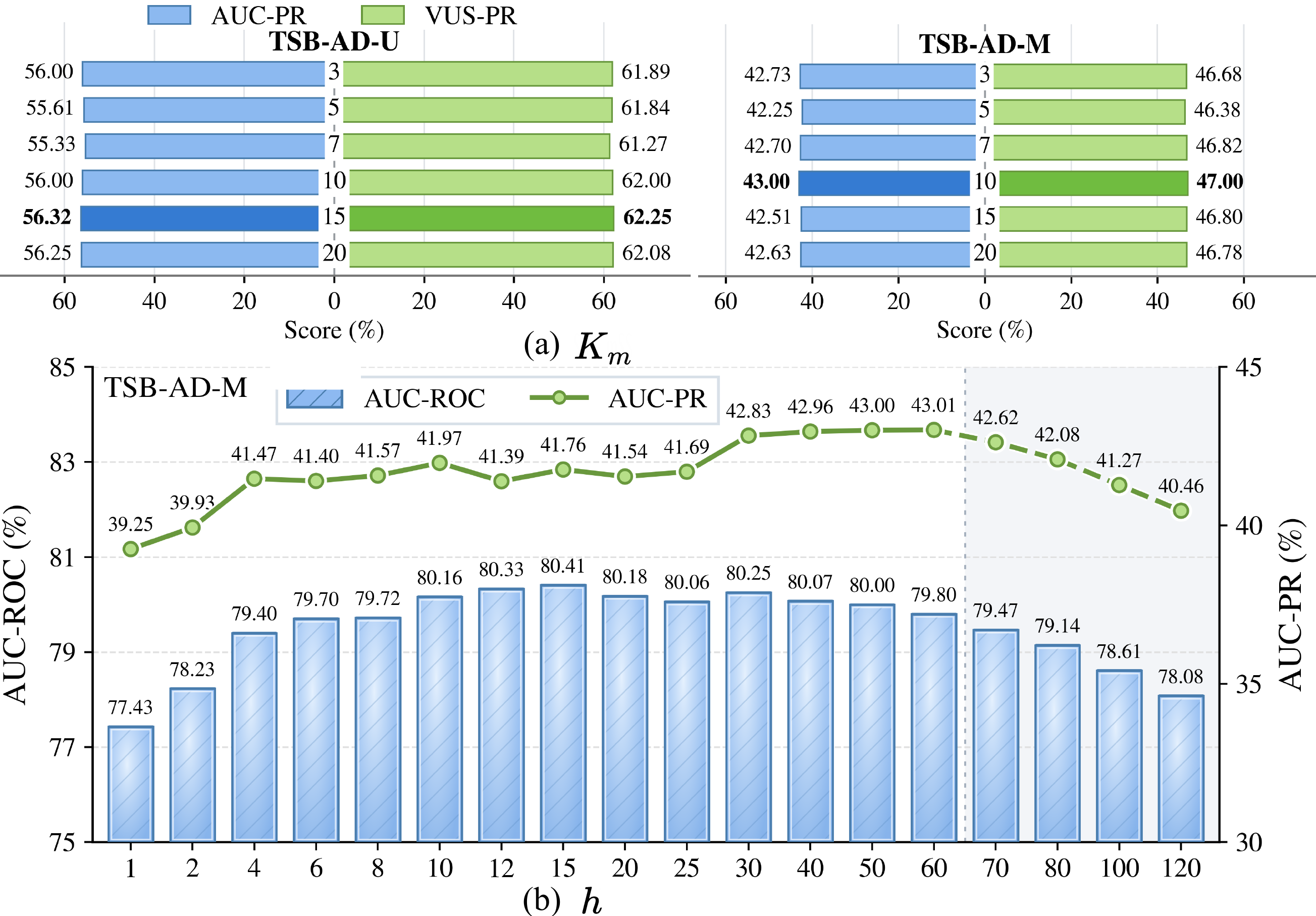}
    \vspace{-0.6cm}
    \caption{Hyperparameter sensitivity analysis of ReFi-TS.
    (a) Effect of the number of normal modes $K_m$.
    (b) Effect of the temporal reference range $h$ on TSB-AD-M.}
    \label{fig:km_h_sensitivity}
\end{figure}

\textbf{\textit{c)  Sensitivity Analysis of $K_m$.}}
Figure~\ref{fig:km_h_sensitivity}(a) investigates the effect
of the number of normal modes $K_m$ on detection performance.
On TSB-AD-U, VUS-PR and AUC-PR reach their highest values
of 62.25\% and 56.32\% at $K_m=15$, respectively.
On TSB-AD-M, both metrics peak at $K_m=10$,
achieving 47.00\% VUS-PR and 43.00\% AUC-PR.
Despite these differences, performance remains relatively
stable across the evaluated settings, with maximum
variations of 0.98 and 0.62 percentage points in VUS-PR
on TSB-AD-U and TSB-AD-M, respectively.
We therefore adopt $K_m=10$ for both datasets,
which achieves the best performance on TSB-AD-M
and remains competitive on TSB-AD-U,
with only 0.25 and 0.32 percentage-point differences
from its optimal VUS-PR and AUC-PR, respectively.


\textbf{\textit{d) Sensitivity Analysis of $h$.}}
Figure~\ref{fig:km_h_sensitivity}(b) illustrates the effect 
of the temporal reference range $h$ on TSB-AD-M.
Increasing $h$ initially improves detection performance 
by providing more representative contextual statistics.
Both AUC-ROC and AUC-PR remain relatively stable across 
intermediate reference ranges, with AUC-PR reaching 
43.00\% at $h=50$.
Further increasing $h$ beyond 60 degrades both metrics, 
suggesting a trade-off between contextual coverage 
and local statistical relevance.
Excessively broad temporal references may mix heterogeneous 
normal modes, weakening the discriminability of contextual 
statistical deviations.

\section{Conclusion}
\label{sec:conclusion}
In this paper, we propose ReFi-TS, a temporal relational 
fingerprint-guided normality learning method for unsupervised 
time-series anomaly detection.
ReFi-TS explicitly characterizes contextual statistical relationships 
and intrinsic temporal structures, leveraging them to guide deep 
representation learning through bidirectional cross-view alignment.
Prototype-guided mode matching coupled with soft mode-conditional  
hypersphere learning enables the characterization of heterogeneous 
normal behaviors.
During inference, deep representation discrepancy, global ReFi rarity, 
and local ReFi disaffinity provide complementary evidence for anomaly detection.
Extensive experiments on TSB-AD-U and TSB-AD-M demonstrate the 
superior performance of ReFi-TS, achieving VUS-PR scores of 
62.00\% and 47.00\%, respectively, with only approximately 
0.4M trainable parameters.


\bibliographystyle{ACM-Reference-Format}
\bibliography{sample}

\end{document}